%% file: acl_latex.tex
\documentclass[11pt]{article}

\usepackage[final]{acl}

\usepackage{times}
\usepackage{latexsym}

\usepackage[T1]{fontenc}

\usepackage[utf8]{inputenc}

\usepackage{microtype}

\usepackage{inconsolata}

\usepackage{graphicx}

\usepackage{hyperref}
\usepackage{amssymb}
\usepackage{url}
\usepackage{amsmath}
\usepackage{algorithm}
\usepackage{algorithmic}
\usepackage{xspace}
\usepackage{xcolor}
\usepackage{enumitem}
\usepackage{tcolorbox}
\usepackage{listings}
\usepackage{pifont}
\usepackage{parskip}

\lstdefinelanguage{json}{
    basicstyle=\ttfamily\small,
    numbers=left,
    numberstyle=\tiny\color{gray},
    stepnumber=1,
    numbersep=6pt,
    showstringspaces=false,
    breaklines=true,
    frame=single,
    literate=
     *{0}{{{\color{blue}0}}}{1}
      {1}{{{\color{blue}1}}}{1}
      {2}{{{\color{blue}2}}}{1}
      {3}{{{\color{blue}3}}}{1}
      {4}{{{\color{blue}4}}}{1}
      {5}{{{\color{blue}5}}}{1}
      {6}{{{\color{blue}6}}}{1}
      {7}{{{\color{blue}7}}}{1}
      {8}{{{\color{blue}8}}}{1}
      {9}{{{\color{blue}9}}}{1}
}
\usepackage[table]{xcolor} 
\definecolor{skyblue}{RGB}{135,206,235}

\usepackage{graphicx}   
\usepackage{subcaption} 
\usepackage{booktabs} 
\usepackage{multirow} 
\usepackage{hhline}   
\usepackage{array} 
\def\modelname{MAGNET\xspace}

\title{MAGNET: Towards Adaptive GUI Agents with Memory-Driven Knowledge Evolution}

\author{
    Libo Sun$^{1}$\thanks{Equal contribution.}, Jiwen Zhang$^{1}$\footnotemark[1], Siyuan Wang$^{3}$, Zhongyu Wei$^{1,2}$\thanks{Corresponding author.}\\
    $^{1}$Fudan University, Shanghai, China\\
    $^{2}$Shanghai Innovation Institute, Shanghai, China\\
    $^{3}$University of Southern California, Los Angeles, USA\\
    \texttt{\{lbsun23,jiwenzhang21\}@m.fudan.edu.cn;sw\_641@usc.edu; zywei@fudan.edu.cn}\\
}
\begin{document}
\maketitle

\begin{abstract}

Mobile GUI agents powered by large foundation models enable autonomous task execution, but frequent updates altering UI appearance and reorganizing workflows cause agents trained on historical data to fail. Despite surface changes, functional semantics and task intents remain fundamentally stable. Building on this insight, we introduce \textbf{\modelname}, a \textbf{m}emory-driven \textbf{a}daptive a\textbf{gent} framework with dual-level memory: \textbf{stationary memory} linking diverse visual features to stable functional semantics for robust action grounding and \textbf{procedural memory} capturing stable task intents across varying workflows. We propose a \textbf{dynamic memory evolution} mechanism continuously refining both memories by prioritizing frequently accessed knowledge. Online benchmark AndroidWorld evaluations show substantial improvements over baselines, while offline benchmarks confirm consistent gains under distribution shifts. These results validate that leveraging stable structures across interface changes improves agent performance and generalization in evolving software environments.
\end{abstract}

\section{Introduction}

Graphical user interfaces (GUIs) are the primary medium for operating mobile devices~\cite{shneiderman2010designing}, yet they require users to specify every low-level action~\cite{hutchins1986direct}, limiting efficiency in complex workflows~\cite{sheridan2005human}. GUI agents powered by multimodal large language models (MLLMs)~\citep{wang2024mobile,liu2024autoglm,wu2024atlas,qin2025ui,wang2025ui} address this by autonomously executing multi-step tasks from natural language instructions. However, ensuring their robustness in continuously evolving software environments remains a critical challenge~\cite{yang2025gui,hu2025agents,sager2025comprehensive}.

\begin{figure}[t]
\setlength{\abovecaptionskip}{0pt}
    \centering
    \includegraphics[width=\linewidth]{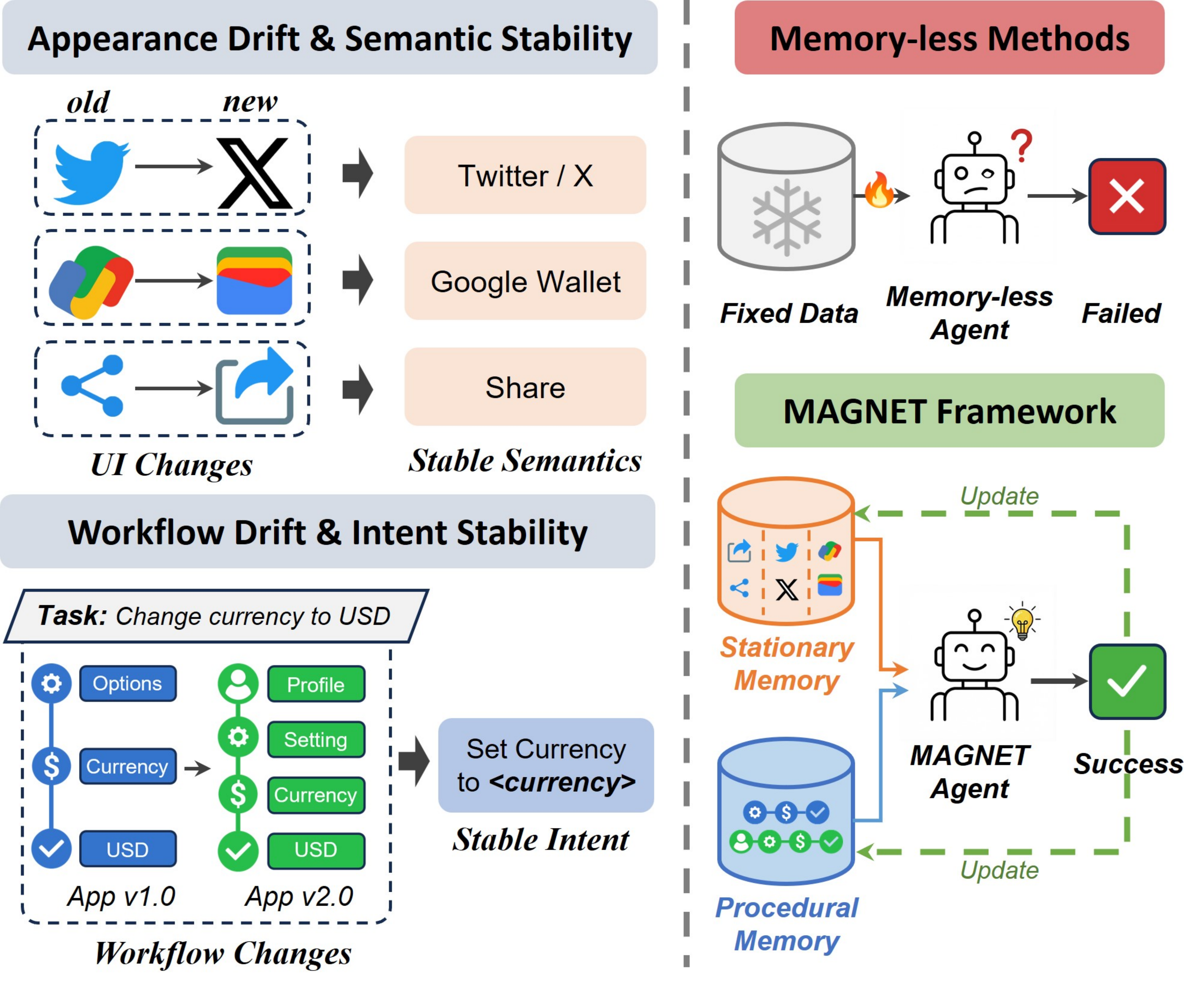}
    \caption{\textbf{Challenges and opportunities in evolving mobile interfaces.} (Left) Two types of drifts 
and their underlying stability: Appearance Drift vs. Semantic 
Stability, Workflow Drift vs. Intent Stability. (Right) MAGNET 
exploits these stable aspects to maintain effectiveness, while memory-less agents relying on frozen knowledge struggle to adapt.
}
    \label{fig:intro_problems}
\end{figure}

\begin{figure*}[t]
\setlength{\abovecaptionskip}{2pt}
    \centering
    \includegraphics[width=0.9\linewidth]{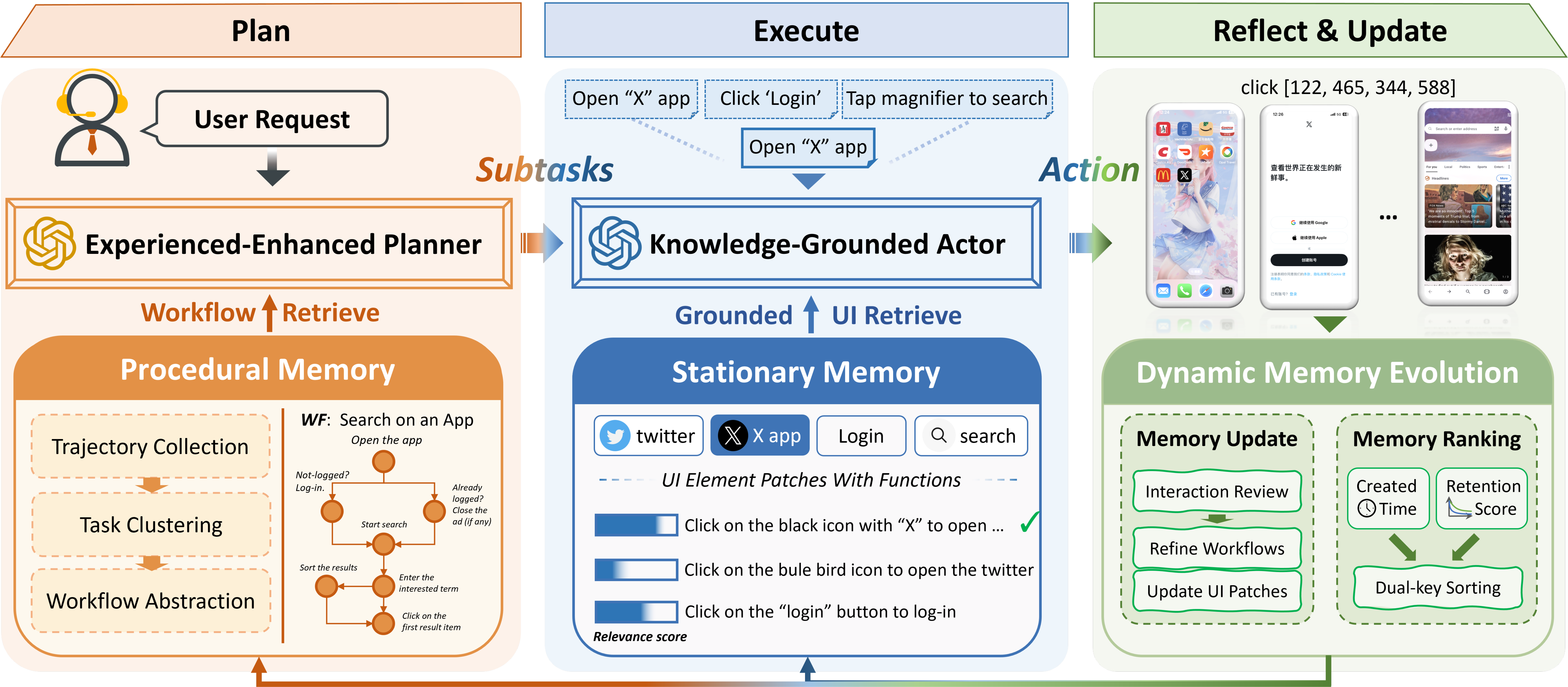}
    \caption{\textbf{\modelname framework.} The planner leverages procedural memory to decompose user requests into subtasks, while the actor grounds each subtask with stationary memory of UI elements.}
    \label{fig:magnet_framework}
\end{figure*}

Mobile ecosystems are inherently dynamic. We find that frequent version updates in commercial applications introduce two forms of drift, as illustrated in Figure~\ref{fig:intro_problems} (left). Firstly, \textit{appearance drift} occurs when UI elements are redesigned without altering their functions (e.g., the icon transition from Twitter to X). Secondly, \textit{workflow drift} arises when operation logic is reorganized across app versions (e.g., the task of "changing the currency to USD"). However, existing specialized models~\citep{hong2024cogagent,zhang2024ui,chen2025guicourse} are trained on fixed datasets that will be outdated as applications evolve, which limits them to generalize to evolving interface states~\cite{wang2024comprehensive}. Although memory-augmented systems~\citep{li2408appagent,agashe2024agent,gao2025chain} attempt to store reusable knowledge, they mainly focus on text-based workflow descriptions that lack multimodal knowledge~\citep{hu2025agents}, making them vulnerable to visual changes in UI elements.

Despite these drifts, we observe that certain properties remain consistent across application updates, as illustrated in Figure~\ref{fig:intro_problems} (left). We summarize these observations as two forms of stability. \textit{Semantic stability} refers to the preservation of functional meaning despite visual redesigns, where updated elements support the same actions. \textit{Intent stability} captures the persistence of high-level task goals even when workflows are reorganized (e.g., changing currency to USD via different navigation paths). These stable properties provide opportunities for building GUI agents that can adapt its actions to interface evolution and continue to complete tasks, as illustrated in Figure~\ref{fig:intro_problems} (right). 

Motivated by this, we introduce a \textbf{m}emory-driven \textbf{a}daptive a\textbf{gent} framework, namely \textbf{\modelname}, that maintains two memory modules to leverage these stabilities. 
Specifically, we use a \textbf{procedural memory} to capture intent stability by storing workflow variants for each task objective, enabling adaptation to reorganized logic. We further propose a \textbf{stationary memory} that captures semantic stability by linking diverse visual features to stable functional semantics for robust grounding under appearance changes.
To support continuous knowledge evolution, we design a \textbf{dynamic memory evolution} mechanism that updates memory and prioritizes frequently accessed information. These updates are extracted by using an automated conrtcution pipeline that distills element–function pairs and workflow templates from completed tasks. 

We evaluate \modelname on three representative offline benchmarks~\citep{zhang2024android,lu2024gui,chai2024amex} to validate the reliability and generalization of the initialized memory. We further assess \modelname in the online AndroidWorld environment~\citep{rawles2024androidworld}, where the memory is continuously updated through interaction to enable adaptation in dynamic settings. Experiments show that MAGNET outperforms zero-shot memory-augmented baselines while remaining competitive with specialized models.

Our primary contributions are:
\begin{itemize}[leftmargin=*, partopsep=0pt, topsep=0pt]
    \setlength{\itemsep}{0pt}
    \setlength{\parsep}{0pt}
    \setlength{\parskip}{0pt}
    \item We identify appearance and workflow drift as key challenges in evolving applications, and design \textbf{\modelname}, a dual-memory framework leveraging semantic and intent stability to address them.
    \item We propose a dynamic memory evolution mechanism together with an automated cnstruction pipeline for continual memory optimization. 
    \item We conduct extensive experiments and validate that \modelname achieves superior adaptability and robust generalization.
\end{itemize}

\section{\modelname Framework}


\modelname (Figure~\ref{fig:magnet_framework}) is a memory-driven agent framework for adaptation in dynamic mobile environments. Following prior planner–actor frameworks~\citep{zheng2024gpt,zhang2024android,gou2024navigating}, we adopt a multi-agent architecture with a \emph{planner} for task decomposition and an \emph{actor} for grounded execution. Distinct from traditional frameworks, \modelname augments the planner with a \textbf{procedural memory} (\S\ref{sec:procedural_mem}) to retrieve abstract workflows, and the actor with a \textbf{stationary memory} (\S\ref{sec:stationary_mem}) to ground actions using UI element exemplars.  Upon receiving a user request, the planner queries procedural memory to retrieve relevant workflows and decomposes the objective into subtasks. For each subtask involving grounding operations, the actor queries stationary memoryto retrieve UI element patches as visual exemplars, enabling reliable action grounding despite interface changes. The actor iteratively executes actions until subtask completion. Finally, we utilize a \textbf{dynamic memory evolution} mechanism (\S\ref{sec:memory_evolution}) to incorporate new workflows and UI elements from successful trajectories while adjusting memory priorities through retention-based ranking, enabling autonomous adaptation to environment updates.

\subsection{Experience-enhanced Planner with Procedural Memory}
\label{sec:procedural_mem}

To mitigate workflow drift caused by application updates, we equip the planner with a procedural memory that stores abstract workflows distilled from completed tasks. Each workflow consists of a task category name (e.g., “Search and Install an App”) and a sequence of high-level steps with categorical placeholders (e.g., [AppName], [SearchQuery]), enabling reuse across different contexts and interface variations.

When receiving a new instruction $I_{\text{new}}$, the planner computes cosine similarity with workflow names in the memory base and retrieves the most similar workflows based on the ranking mechanism (\S\ref{sec:memory_ranking}). The retrieved workflows, including their names and step sequences, are concatenated into the planner's context, which comprises the current screenshot, instruction, screen description, and historical actions. By integrating these workflows, the planner generates subtasks that decompose the high-level objective into executable steps.

\paragraph{Memory Construction.}
To initialize the procedural memory, we propose a three-stage automatic construction pipeline, detailed as follows: 

\begin{itemize}[leftmargin=*, partopsep=0pt, topsep=0pt]
    \setlength{\itemsep}{0pt}
    \setlength{\parsep}{0pt}
    \setlength{\parskip}{0pt}
    \item \textbf{Trajectory Collection}: We collect instruction-trajectory pairs $\{(I_j, \pi_j)\}_{j=1}^M$ from human demonstrations or curated datasets, where $\pi_j = \langle a_1^j, \ldots, a_{k_j}^j \rangle$ represents action sequences that successfully complete tasks. 
    \item \textbf{Task Clustering}: To group tasks with similar workflows, we embed instructions using MiniCPM-Embedding~\citep{hu2024minicpm} and construct a similarity graph $\mathcal{G}$ by connecting instruction pairs whose cosine similarity exceeds a threshold $\tau$. We then extract maximal cliques $\mathcal{C} = \{c_1, c_2, \ldots, c_m\}$~\citep{sun2019using} as task clusters (Algorithm~\ref{clustering_algorithm}), where each cluster $c_i$ groups instructions with similar task patterns.
    \item \textbf{Workflow Abstraction}: For each cluster $c_i \in \mathcal{C}$, associated trajectories are synthesized into abstract workflows by prompting the planner to distill common patterns~\citep{wu2024copilot,yang2024buffer} (Appendix~\ref{memory_construction}).

\end{itemize}



\subsection{Knowledge-grounded Actor with Stationary Memory}
\label{sec:stationary_mem}

To address appearance drift caused by UI updates, we ground the actor with a stationary memory that associates visual representations of UI elements with their functional intents. The memory stores pairs $\langle d_i, v_i \rangle$, where $d_i$ describes an element’s function (e.g., “click the search icon to start searching”) and $v_i$ is the corresponding visual patch. This representation enables the actor to generalize across interface variations by reasoning about functionality rather than exact visual appearance.


When the planner generates subtasks involving UI element localization, the subtask description is used as a query to compute cosine similarity with functional descriptions $d_i$ in the memory base. The most similar entries are retrieved using the ranking mechanism (\S\ref{sec:memory_ranking}), and their visual patches $v_i$ are extracted as references. For grounding based on general MLLMs, these patches are concatenated into the actor's context (screenshot, instruction, description, historical actions, subtask), enabling robust identification of target UI elements despite appearance variations and interface changes. 
For specialized grounding models with fixed input formats, stationary memory is injected via a lightweight adaptation: retrieved patches are template-matched to the current screenshot to locate a center point, around which a bounding box enclosing the nearest $k$ icons is drawn as a visual hint. This introduces stationary memory without modifying the model’s input format, demonstrating the adaptability of our stationary memory.

\begin{figure*}[t]
\setlength{\abovecaptionskip}{0pt}
    \centering
    \includegraphics[width=\linewidth]{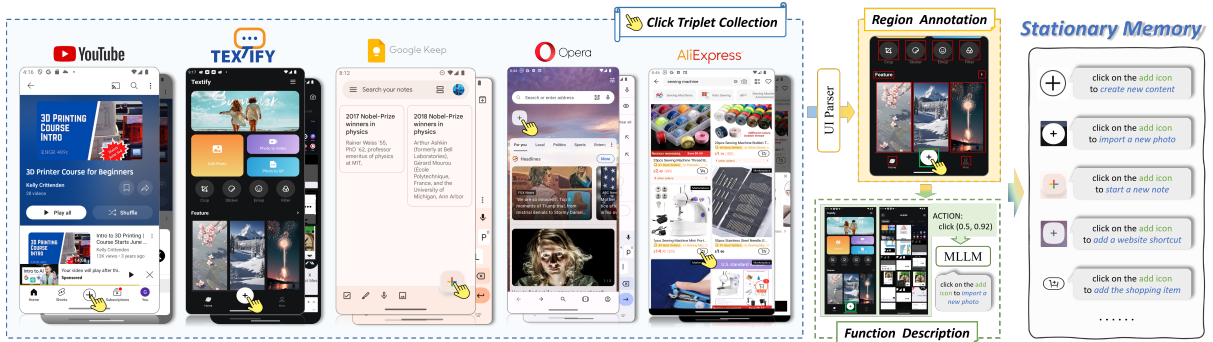}
    \caption{\textbf{Construction pipeline of stationary memory.} Each newly generated entry $\langle d_i, v_i \rangle$ is checked against the stationary memory by retrieving similar functional descriptions and corresponding UI element patches. If a duplicate entry is detected, it is discarded to avoid redundancy.}
    \label{fig:pipeline_stationary}
\end{figure*}

\paragraph{Memory Construction.}
To initialize the stationary memory, we design an automatic construction pipeline composed of three stages illustrated in Figure~\ref{fig:pipeline_stationary}. Specifically, it involves: 

\begin{itemize}[leftmargin=*, partopsep=0pt, topsep=0pt]
    \setlength{\itemsep}{0pt}
    \setlength{\parsep}{0pt}
    \setlength{\parskip}{0pt}
    \item \textbf{Triplet Collection}: Screen-action-screen triplets $\langle o_{t}, a_t, o_{t+1} \rangle$ are extracted from various sources including offline episodic GUI datasets and online interaction trajectories. Each triplet comprises consecutive screenshots captured before and after a click action, enabling inference of both the visual target and functional outcome.
    \item \textbf{Region Annotation}: We employ OmniParserV2~\citep{yu2025omniparser} to parse screenshots into candidate regions with bounding boxes. The clicked element is identified by selecting the candidate region whose spatial layout best aligns with the click position, and the resulting bounding box crops the element patch $v_k$.
    \item \textbf{Function Description}: We apply Qwen2.5-VL-32B~\citep{bai2025qwen2} to infer functional semantics $d_i$ in a consistent format (Appendix~\ref{memory_construction}). This ensures each visual patch $v_i$ is consistently paired with its functional intention $d_i$.
\end{itemize}


\paragraph{The UI-40K Dataset.}
Applying this pipeline to AITZ~\citep{zhang2024android}, GUI-Odyssey~\citep{lu2024gui}, and Amex~\citep{chai2024amex}, we construct the \textbf{UI-40K} dataset containing 41,009 multimodal entries $\langle o_t, a_t, d_t, o_{t+1} \rangle$ across 20,618 unique functional intents. Beyond serving as stationary memory in \modelname, UI-40K has broader potential applications including GUI grounding for instruction-tuning and offline reinforcement learning. We confirm UI-40K with high quality through both automatic validation using SOTA MLLMs and manual verification (detailed in Appendix~\ref{appendix:ui40k_validation}).

\subsection{Dynamic Memory Evolution}
\label{sec:memory_evolution}

To ensure memories remain effective as applications evolve, both memory modules employ content update and adaptive ranking mechanisms.

\subsubsection{Memory Content Update}
\label{sec:memory_update}


\paragraph{Procedural Memory}
When successful task trajectories are obtained, the workflow abstraction process (\S\ref{sec:procedural_mem}) extracts structured workflows. Each extracted workflow is stored as an independent memory entry with a creation timestamp $\tau_i$ and retrieval count $n_i=0$, preserving multiple execution paths for similar tasks. During retrieval, the ranking mechanism (\S\ref{sec:memory_ranking}) balances usage frequency and recency to prioritize reliable workflows while down-weighting outdated ones, enabling procedural memory to adapt to interface changes.

\paragraph{Stationary Memory}
When new grounding actions are observed, the system constructs $\langle d_i, v_i \rangle$ pairs via the \S\ref{sec:stationary_mem} pipeline. For each pair, the system queries for matching functional descriptions. If found, the new visual patch $v_i$ is appended to the existing entry's image list with independent timestamp $\tau_i$ and retrieval metadata ($n_i, t_i$), consolidating visual variants under shared functions where each image independently tracks its usage history. If no match exists, a new entry is created. During retrieval, images are ranked by individual retention scores and updated time.

\input{main_result}

\subsubsection{Memory Ranking for Retrieval}
\label{sec:memory_ranking}


Memory retrieval in dynamic environments must balance contextual relevance with avoiding outdated knowledge as interfaces evolve. Therefore, we introduce a memory ranking mechanism that prioritizes reliable entries by considering two signals: knowledge creation time (timestamp $\tau_i$), which favors entries aligned with current application versions, and usage history, which reflects empirical utility through retrieval frequency and recency. To quantify usage history, the system maintains a global counter $C_{\text{global}}$ and per-entry metadata (last access time $t_i$ and total retrieval count $n_i$), from which an inactivity gap $g_i = C_{\text{global}} - t_i$ is computed. Inspired by the Ebbinghaus forgetting curve~\citep{ebbinghaus1885gedachtnis}, we derive a retention score $R_i = \exp(-g_i / n_i)$, where frequently accessed entries decay more slowly.

Based on these signals, we adopt a two-stage retrieval strategy: the first stage uses semantic similarity to select the top-$N$ candidates relevant to the query; the second stage ranks these candidates using the retention score $R_i$ and creation timestamp $\tau_i$ to produce the final top-$K$ entries. Complete algorithmic details are provided in Algorithm~\ref{alg:memory_evolution}.

\section{Experiments}

\subsection{Experimental Setup}
\label{subsec:exp_setup}

\paragraph{Evaluation Benchmarks} 

We evaluate \modelname on both offline and online settings.
We conduct offline evaluations using AITZ~\citep{zhang2024android}, GUI-Odyssey~\citep{lu2024gui}, and Amex~\citep{chai2024amex}. To assess distribution shift generalization, we create custom splits for GUI-Odyssey and Amex, yielding in-distribution (ID), template-shifted (TS), app-shifted (AS), and domain-shifted (DS) subsets (Details in Appendix~\ref{appendix:data}). 
Building on the validated offline results, we further evaluate \modelname in an online environment AndroidWorld~\citep{rawles2024androidworld}. Following the official protocol, all methods share the same action budget and public exploration scripts. 
Detailed settings are provided in Appendix~\ref{appendix:online_settings}.

\paragraph{Baselines}
We evaluate \modelname against two categories of baselines.
For offline evaluation, we compare with end-to-end specialized models, including UI-Venus-Navi-7B~\citep{gu2025ui}, Atlas-Pro-7B~\citep{wu2024atlas}, GUI-R1-7B~\citep{luo2025gui}, and InfiGUI-R1-3B~\citep{liu2025infigui}, as well as the memory-free COAT framework~\citep{zhang2024android} and the memory-augmented Agent-S~\citep{agashe2024agent}. We evaluate \modelname with two representative backbones, Qwen2.5-VL-32B and Gemini-2.5-Pro, covering both open- and closed-source models with different capability levels. Detailed configurations are provided in Appendix~\ref{appendix:offline_baselines}.
For {online evaluation}, we compare \modelname with M3A~\citep{rawles2024androidworld}, memory-augmented AppAgent~\citep{zhang2025appagent}, and Agent-S~\citep{agashe2024agent}. All online agents use Qwen2.5-VL-32B as the backbone~\citep{bai2025qwen2}.
In the online evaluation, memory-augmented agents first perform self-exploration or task execution to initialize their memory, and then leverage the accumulated memory during the evaluation phase. Implementation details are provided in Appendix~\ref{appendix:online_baselines}.

\setlength{\abovecaptionskip}{2pt}
\begin{table}[t]
\centering
\resizebox{0.95\linewidth}{!}{
\begin{tabular}{lc}
\toprule
\textbf{Method} & \textbf{SR (\%)} \\
\midrule
M3A~\citep{rawles2024androidworld} & 32.78 \\
AppAgent~\citep{zhang2025appagent} & 34.43 \\
Agent-S~\citep{agashe2024agent} & 40.98 \\
\modelname{} (Ours) & \textbf{42.62} \\
\bottomrule
\end{tabular}
}
\caption{\textbf{Comparison with other memory-augmented baselines on the online AndroidWorld environment.}}
\label{tab:androidworld_baseline}
\end{table}

\paragraph{Evaluation Metrics}
For {offline evaluation}, we use two step-level metrics: Success Rate (SR), which measures the accuracy of predicted actions at each step, and Grounding accuracy (Grd.), which measures the accuracy of coordinate prediction for click actions. Details are in Appendix~\ref{appendix:eval_metric}. For {online evaluation}, we measure task completion rate (SR) using AndroidWorld's official evaluation scripts, which assess whether agents successfully complete end-to-end tasks in the live environment. 


\subsection{Main Results}

\input{ablation_iid_new}

\paragraph{Offline Evaluation}
Table~\ref{tab:main_results} reports results on the in-distribution subsets across three datasets. 
With Qwen2.5-VL-32B as backbone, \modelname consistently outperforms COAT and Agent-S across all three datasets, demonstrating the effectiveness of the proposed dual-memory design under a fixed foundation model. 
While specialized models achieve higher absolute scores, \modelname with Gemini-2.5-Pro as backbone further improves performance  without task-specific training 
, indicating that the proposed framework can effectively benefit from stronger foundation models. 

\paragraph{Online Evaluation}
Table~\ref{tab:androidworld_baseline} presents online evaluation results on AndroidWorld. All agents employ self-exploration to initialize their memory using three task execution episodes (detailed in Appendix~\ref{appendix:online_baselines}). AppAgent stores UI elements and their functions using element identifiers from XML page structures, while Agent-S emphasizes workflow-level memory but lacks grounded UI representations.
In contrast, \modelname integrates stationary memory with multimodal grounding of UI elements and procedural memory with reusable workflows, enabling the agent to reason jointly over interface structure and action-level experience. As a result, \modelname achieves a 42.62\% task completion rate, outperforming AppAgent (34.43\%) by +8.2 points and Agent-S (40.98\%) by +1.6 points. This performance gap underscores the importance of decoupling memory representations from page-specific identifiers while jointly modeling UI-level and workflow-level knowledge for robust task completion in dynamic environments.

\input{cross_base_results_concise}

\subsection{Ablation Study}
\paragraph{Memory component effectiveness.}
Table~\ref{tab:ablation_iid} isolates the effects of each memory component on ID subsets. We evaluate four configurations: baseline, procedural only, stationary only, and both combined. Results show both memories provide consistent gains, with their combination achieving optimal performance. Procedural memory yields larger SR improvements (+2.66\% on AMEX with Qwen) than stationary memory (+0.52\%), because stationary memory mainly benefits cases in which models lack prior coverage of specific UI icons, a condition that is underrepresented in current static offline benchmarks. However, stationary memory shows clear value in Grd improvements (ranging from +0.26\% to +0.88\%) and in cases with novel icons (detailed in Appendix~\ref{appendix:case_studies}).

\paragraph{Robustness across architectures.}
Table~\ref{tab:paired_ablation} demonstrates the average effectiveness of \modelname among three datasets (detailed in \ref{appendix:detailed_ablation}) across five planner–actor configurations, categorized into \textit{Homogeneous} (same model for planner and actor) and \textit{Heterogeneous} (different models) types. \modelname consistently improves performance with average gains of +2.3\% SR and +2.7\% Grd across all configurations. Improvements are particularly pronounced with heterogeneous pairings (e.g., QwenVL + OS-Atlas: +4.2\% SR), where the actor is a specialized grounding model that incorporates retrieved stationary memory using the injection strategy described in \S\ref{sec:stationary_mem}, suggesting that memory can effectively compensate for limitations or mismatches between planning and execution modules by providing reusable intermediate knowledge. Even with strong homogeneous setups, memory still offers clear benefits (Gemini + Gemini: +1.5\% SR), indicating that its gains are not limited to weaker backbones. Overall, these results demonstrate that the proposed memory design generalizes across architectural choices and supports flexible planner–actor deployment.

\begin{figure}[t]
\setlength{\abovecaptionskip}{-5pt}
    \centering
    \includegraphics[width=\linewidth]{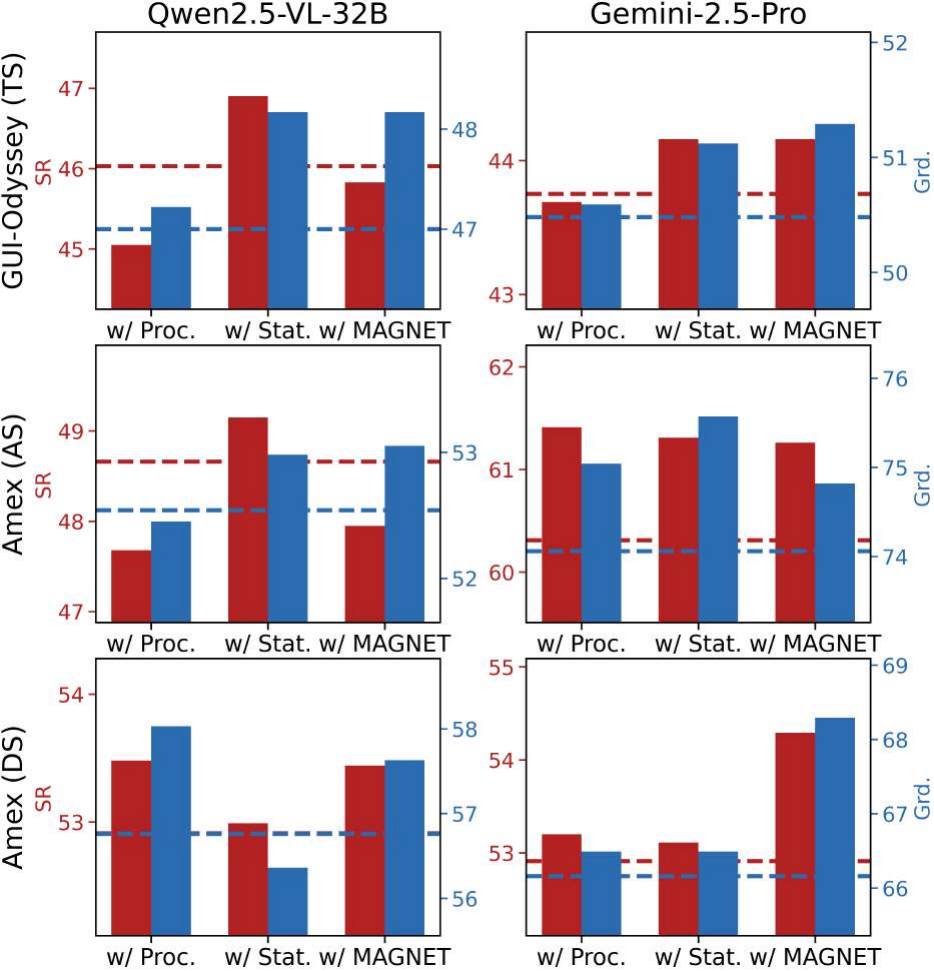}
    \vspace{5pt}
    \caption{
    \textbf{Results on the template, app, and domain shifted subsets.} The MLLM serves as both the planner and the actor. The dashed lines in the figure indicate the baseline performance.
    }
    \label{fig:ablation_ood}
\end{figure}

\subsection{Generalization Discussion}

Figure~\ref{fig:ablation_ood} evaluates generalization on template-shifted, app-shifted, and domain-shifted subsets, where agents encounter scenarios beyond source memory coverage. Despite operating under distribution shifts, memory-augmented agents outperform the baseline across most settings, indicating that external memory remains beneficial beyond training coverage.
With Qwen2.5-VL-32B, stationary memory leads to modest but consistent SR improvements (+0.1\% to +0.9\%) across most shift types, while procedural memory exhibits higher variance, ranging from slight degradation under template- and app-shifted settings to moderate gains under domain shifts. In contrast, Gemini-2.5-Pro demonstrates more stable improvements from both memory types (+0.2\% to +1.4\% SR), indicating a stronger ability to integrate heterogeneous memory signals under distribution shifts.
Overall, these trends suggest that different distribution shifts emphasize different forms of memory information: stationary memory being more effective when surface-level structures remain informative, and procedural memory becoming more critical when task objectives and execution patterns change. Moreover, the reduced sensitivity to shift type for Gemini-2.5-Pro suggests that stronger models can more flexibly integrate multiple memory signals. Together, these results provide empirical evidence that our memory design captures complementary aspects of past experience, enabling agents to adapt across diverse generalization scenarios.

\begin{table}[t]
\small
\centering
\captionsetup{skip=2pt}
\setlength{\tabcolsep}{2pt}
\setlength{\belowcaptionskip}{3pt}
\begin{tabular}{lccc}
\toprule
\textbf{Iteration}  & \textbf{SR (\%)} & \textbf{Proc$_{\text{Amex}}$ (\%)} & \textbf{Stat$_{\text{Amex}}$ (\%)} \\
\midrule
\modelname{} & 31.14 & 100 & 100 \\
1 iteration  & 37.70 & 38 & 27 \\
2 iterations & 39.34 & 32 & 24 \\
3 iterations & 40.98 & 26 & 18 \\
\bottomrule
\end{tabular}
\caption{\textbf{In-the-wild memory evolution on AndroidWorld. }
Proc$_{\text{Amex}}$ and Stat$_{\text{Amex}}$ denote the percentage of retrieved procedural and stationary memories originating from the initial Amex memory bank. 
}
\label{tab:androidworld_iterations}
\end{table}

\subsection{Continual Adaption Discussion}

To evaluate \modelname's capacity for continual learning, we warm-start the agent with memories from Amex and then iteratively deploy it on AndroidWorld tasks. After each iteration, the newly acquired interaction trajectories and grounding experiences are incorporated into the memory base and used to initialize the agent for the subsequent iteration. Table~\ref{tab:androidworld_iterations} shows success rate rising from 31.14\% to 40.98\% over three iterations. The Proc$_{Amex}$ and Stat$_{Amex}$ columns measure the percentage of retrieved memory entries that still originate from the initial Amex bank (procedural and stationary, respectively). Both proportions decrease sharply, from 100\% to 26\% for procedural memory and from 100\% to 18\% for stationary memory, indicating that Amex-derived knowledge is progressively replaced by AndroidWorld-specific experiences. This trend confirms that dynamic memory update mechanism effectively discards outdated or less relevant memories and adapts to new environments through continuous online learning.


\section{Related Works}

\subsection{MLLM Agents}

The emergent capabilities of Multimodal Large Language Models (MLLMs) have motivated their use as central controllers that orchestrate external components~\citep{wu2023visual, li2023modelscope, yang2023mm, wang2024genartist, shen2024small}. These agents augment MLLMs with memory~\citep{fan2024videoagent, wang2024jarvis}, tool use~\citep{schick2023toolformer, wang2025mllm}, complex reasoning~\citep{yang2023mm, wang2024genartist} and the ability of iterative learning in real environments~\citep{qian2024iterative, xi2024agentgym}.  Currently, MLLM-driven agents are flourishing across a broad spectrum of applications, ranging from general-purpose tasks such as image generation and editing~\citep{wang2024genartist} and video games~\citep{wang2023voyager,li2025optimus}, to domain-specific areas including healthcare~\citep{li2024mmedagent} and e-commerce~\citep{gong2025mindflow}. Building on these advances, our work extends to perceive and operate GUI within dynamic mobile environments.

\subsection{GUI Agents}

As a specialized instantiation of MLLM agents, GUI agents focus on controlling software interfaces and operating systems.
One line of work aims to construct specialist end-to-end agents~\citep{hong2024cogagent, xu2024aguvis, zhang2024ui} by fine-tuning small open-source MLLMs on task specific GUI datasets~\citep{deng2023mind2web, zhang2024android,lu2024gui}. 
Such methods achieve strong in-domain performance with efficient inference, but their heavy reliance on the high-quality labeled datasets~\citep{chen2025guicourse} severely constrains their generalization ability on unseen applications.
Another line of research, including AppAgent~\citep{zhang2025appagent, li2408appagent, jiang2025appagentx} and MobileAgent~\citep{wang2024mobile,wang2024mobilev2,wang2025mobile,qin2025ui}, adopts a planner–actor decomposition, where a planner leverages strong proprietary models (e.g., GPT-4o~\citep{openai_gpt4o}, Gemini~\citep{google_gemini_2_5}) to derive operation steps, and an actor executes actions on screens. These frameworks typically assume actors with strong grounding capabilities, such as SeeClick~\citep{cheng2024seeclick}, UGround~\citep{gou2024navigating}, and OS-Atlas~\citep{wu2024atlas}. While adaptable and interpretable, they require careful system design. OS-Copilot~\citep{wu2024copilot} further extends this paradigm to OS-level agents by unifying heterogeneous system control, but its text-centric self-directed learning offers limited support for visual interface modeling. In contrast, MAGNET targets GUI agents with UI-centric actor memory and workflow-generalized planning, enabling robust adaptation in dynamic interfaces.

\subsection{Memory-enhanced Agents}

LLM agents increasingly incorporate memory to support long-term and complex reasoning. Early work~\citep{park2023generative} introduced natural language memory streams for experience recording and reflection, followed by richer memory designs, including summarization and forgetting~\citep{zhong2024memorybank}, OS-inspired virtual memory management~\citep{packer2023memgpt}, and triplet-based or neurosymbolic representations for precise reasoning~\citep{modarressi2023ret, wang2024symbolic}.
Hierarchical and dynamic memory further improve efficiency in long-horizon tasks, exemplified by subgoal summarization in \citep{hu2024hiagent} and evolving note-like memory in \citep{xu2025mem}. 
At a higher level, memory has been extended toward reusable reasoning patterns and skills: \citep{yang2024buffer} studied buffered reasoning strategies for iterative problem solving, and \citep{wang2023voyager} demonstrated expandable skill libraries that support continual learning in interactive environments. 
More recently, Chain-of-Memory~\citep{gao2025chain} proposed modular memory organization for cross-application navigation and knowledge reuse.
Inspired by these, we introduce a memory-driven framework \modelname to evolve procedural and stationary knowledge about dynamic environments.

\section{Conclusions}

We introduce \modelname, addressing the appearance drift and workflow drift challenges in evolving applications through dual-level memory: stationary memory enables robust grounding despite interface changes, while procedural memory adapts to workflow evolution. A dynamic update mechanism refines memories by prioritizing frequently accessed knowledge. Comprehensive evaluations on AndroidWorld and offline benchmarks demonstrate effectiveness across diverse scenarios, with ablation studies confirming complementary benefits of both components. This work establishes that exploiting semantic and intent stability beneath surface changes enables practical deployment of adaptive agents in evolving software ecosystems.



\section*{Limitations}

While \modelname demonstrates robust adaptation, it has limitations. The framework requires successful trajectories for memory construction, making it less effective in completely novel domains where initial exploration fails. Additionally, the clustering-based workflow extraction may struggle with highly diverse task structures that do not form clear patterns. Future work could explore zero-shot memory initialization and more flexible workflow representations.



\bibliography{custom}

\clearpage
\appendix

\section{Validation of UI-40K Annotation Quality}
\label{appendix:ui40k_validation}

To ensure the reliability of UI-40K annotations, we employed a hybrid validation approach combining automated SOTA MLLM prediction and manual verification.

\paragraph{Automatic Validation.}
We sampled 1,000 instances and queried two state-of-the-art MLLMs (GPT-4o and Gemini-2.5-Pro) to independently generate bounding boxes based on the tuple $\langle$Current Screen, Click Description$\rangle$. We calculated the IoU between the outputs of these two models to filter for consensus. Instances where the inter-model IoU exceeded 0.75 were treated as high-confidence pseudo-ground truth (787 instances), while the remaining 213 instances were assigned to a manual verification set. For the 787 high-confidence instances, we computed the IoU against the OmniParserV2 annotations:

\begin{table}[h]
\centering
\begin{tabular}{ll}
\toprule
\textbf{Metric} & \textbf{Value} \\
\midrule
Mean IoU & 0.55 \\
Median IoU & 0.63 \\
IoU $\geq$ 0.5 & 68.5\% \\
IoU $\geq$ 0.7 & 42.3\% \\
\bottomrule
\end{tabular}
\caption{Automatic validation metrics for UI-40K annotations.}
\label{tab:ui40k_auto_validation}
\end{table}

\paragraph{Analysis of Discrepancies.}
Although the mean IoU is 0.55, manual inspection reveals that discrepancies primarily arise from valid variations in annotation granularity (e.g., the parser marks a larger clickable container while the model marks the specific icon inside), rather than functional errors.

\paragraph{Manual Verification.}
For the remaining 213 cases where models disagreed, manual review confirmed that the annotated regions were semantically correct regarding the clickable element. This validation confirms the high reliability of UI-40K for training and evaluation purposes.

\label{appendix:experiment_detail}

\section{Datasets}
\label{appendix:data}

\subsection{Dataset Splitting}
\label{dataset_splitting}
\paragraph{AITZ} 
The AITZ~\citep{zhang2024android} dataset comes pre-split. We directly use the training set as the \textbf{source} data for memory construction and the test set as the \textbf{in-distribution (ID)} validation set.
\paragraph{GUI-Odyssey} 

The GUI-Odyssey dataset constructs instructions by first generating tasks from templates and then rephrasing them into natural language.  
To create ID and TS splits, we proceed in two steps:

\begin{itemize}[leftmargin=*, partopsep=0pt, topsep=0pt]
    \setlength{\itemsep}{0pt}
    \setlength{\parsep}{0pt}
    \setlength{\parskip}{0pt}
    \item \textbf{Template-level split.} We randomly sample 20\% of all templates, and treat the corresponding data as \textbf{template-shifted (TS)}.
    \item \textbf{Instruction-level split.} For the rest portion, we further divide instructions: 75\% are used as the \textbf{source} data for memory construction, while the remaining 25\% form the \textbf{in-distribution (ID)} test set.  
\end{itemize}

This design ensures that all TS instructions originate from unseen templates, while ID instructions come from seen templates.  
Given the large scale of GUI-Odyssey (over 110k images), we down-sample the final evaluation sets to 309 ID samples and 311 TS samples, yielding a representative size comparable to other datasets.

\paragraph{Amex} 
The Amex dataset instructions also follow certain templates, similar to GUI-Odyssey, although this is not explicitly stated in the original paper or dataset. To address this, we manually classified the instructions, identifying 58 apps and grouping them into 12 domains. 

The Amex splitting strategy is defined as follows:
\begin{enumerate}[leftmargin=0.5cm]
\item \textbf{Domain-level split:} We select three domains (Sports \& Fitness, Others, and Weather) as unseen domains, and the episodes from these domains form the \textbf{domain-shifted (DS)} subset.
\item \textbf{App-level split:} From each of the remaining domains, we randomly choose one app to form the unseen-apps set, while the rest constitute the seen-apps set. Although unseen-apps are disjoint from seen-apps, they still belong to domains already represented in seen-apps; therefore, their data is referred to as the \textbf{app-shifted (AS)} subset.
\item \textbf{Instruction-level split:} For each app in the seen-apps set, we randomly partition instructions into \textbf{source} and \textbf{in-distribution (ID)} subsets using a 4:1 ratio.
\end{enumerate}

\begin{figure*}[h]
    \centering
    \includegraphics[width=0.8\linewidth]{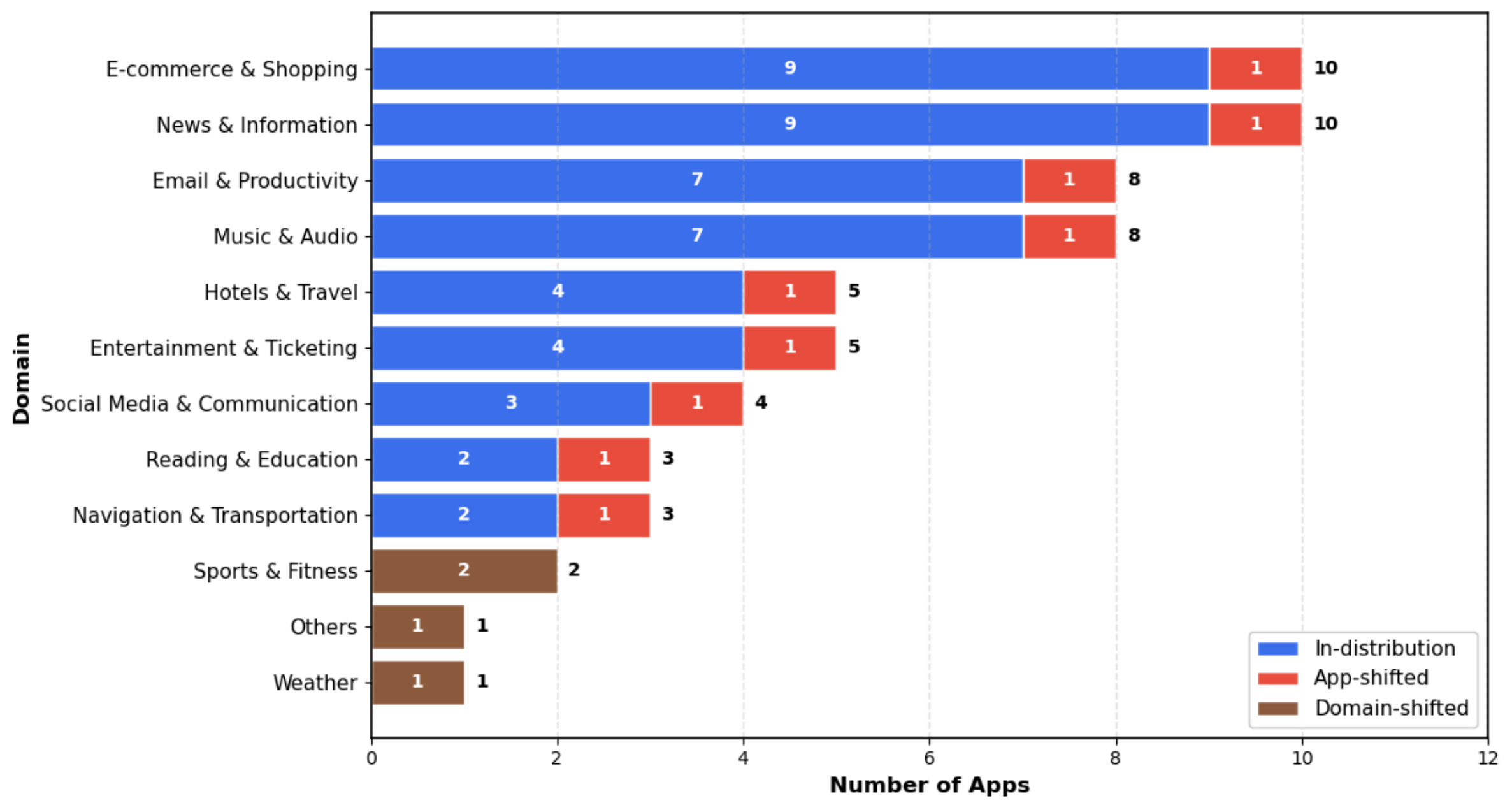}
    \caption{Domain/App distribution of the Amex dataset.}
    \label{fig:amex_domain_app}
\end{figure*}

This yields four subsets in total: source, ID, AS, and DS. The distribution is summarized in Figure~\ref{fig:amex_domain_app}.

\subsection{Statistics}

Table~\ref{tab:dataset_stats} summarizes the statistics of all datasets and their subsets.

\begin{table}[h]
\centering
\begin{tabular}{llrr}
\toprule[1.2pt]
\textbf{Dataset} & \textbf{Subset} & \textbf{Episodes} & \textbf{Screens} \\
\midrule
\multirow{2}{*}{AITZ}    & source        & 1,998 & 13,919 \\
        & in-domain            &   506 &  4,724 \\
\midrule
\multirow{3}{*}{Odyssey} & source        & 4,635 & 71,803 \\
        & in-domain            &   309 &  4,920 \\
        & template-shifted            &   311 &  4,386 \\
\midrule
\multirow{4}{*}{Amex}    & source        & 1,794 & 22,639 \\
        & in-domain            &   448 &  5,563 \\
        & app-shifted            &   487 &  8,035 \\
        & domain-shifted            &   259 &  2,472 \\
\bottomrule[1.2pt]
\end{tabular}
\caption{Dataset Statistics}
\label{tab:dataset_stats}
\end{table}



\section{Experiment Details}

\subsection{Online Agent Configurations}
\label{appendix:online_settings}

\paragraph{Choice of Easy Split.}
AndroidWorld~\citep{rawles2024androidworld} provides three difficulty levels: Easy (61 tasks), Medium (36 tasks), and Hard (19 tasks). We evaluate on the Easy split for the following reasons:

\begin{enumerate}[leftmargin=0.5cm]    
    \item \textbf{Iterative Learning Requirements}: Our continual learning approach requires multiple training iterations (three memory-free passes) to evolve memories from online experiences. The Easy split provides sufficient task diversity while maintaining experimental tractability for this iterative process.
    
    \item \textbf{Research Focus}: Our primary contribution is the memory-driven adaptation framework demonstrated through comprehensive offline benchmarks (AITZ, GUI-Odyssey, Amex with multiple distribution shifts). The online evaluation serves to validate real-world applicability rather than benchmark performance optimization.
\end{enumerate}

\subsection{Baselines Details}

\subsubsection{Offline Baselines}
\label{appendix:offline_baselines}

For offline evaluation, we compare \modelname with both specialized models and agentic frameworks.

\textbf{Specialized models} including UI-Venus-Navi-7B~\cite{gu2025ui}, Atlas-Pro-7B~\cite{wu2024atlas}, GUI-R1-7B~\cite{luo2025gui}, and InfiGUI-R1-3B~\cite{liu2025infigui} are implemented following their respective papers, including model architectures, input formats, and inference settings. For fair comparison across datasets, we unify the action space while preserving each model’s original design assumptions.

\textbf{COAT}~\citep{zhang2024android} is evaluated using the Gemini-2.5-Pro backbone, consistent with its original implementation. COAT does not employ explicit memory mechanisms and serves as a representative memory-free agentic baseline.

\textbf{Agent-S}~\citep{agashe2024agent} is adapted to offline datasets following its original framework design. Memory construction is performed using demonstrations from the corresponding offline dataset, consistent with the experience-augmented retrieval mechanism described in the Agent-S paper. Since offline benchmarks do not support interactive self-exploration, the Episodic Memory component—which relies on online trajectory collection—is omitted. Additionally, the web knowledge module is removed to ensure a fair comparison in a closed-world offline setting. All other components, including hierarchical planning and narrative memory retrieval, are retained without modification.

Unless otherwise specified, all agentic frameworks are evaluated under the same action space.

\subsubsection{Online Baselines}
\label{appendix:online_baselines}
For completeness, we document how each online baseline prepares its memory before evaluation on the AndroidWorld Easy split:
\begin{itemize}[leftmargin=*, partopsep=0pt, topsep=0pt]
    \setlength{\itemsep}{0pt}
    \setlength{\parsep}{0pt}
    \setlength{\parskip}{0pt}
    \item \textbf{AppAgent}~\citep{zhang2025appagent}: For each of the 24 applications contained in the Easy set, AppAgent is paired with a manual exploration task. It runs the task once to produce structured documentation for that application and then reuses the resulting documentation during inference on the downstream tasks.
    \item \textbf{Agent-S}~\citep{agashe2024agent}: The agent first executes one round of generic tasks together with the official exploration tasks to build its internal memory. This single-run memory is then reused for evaluation without additional updates.
    \item \textbf{\modelname}: We run three memory-free passes across the 61 tasks to harvest trajectories, filter them through the metabolism rule, and then conduct the reported evaluation. For the continual-learning study, the initial memories are constructed from the Amex dataset and updated after each full pass over the Easy tasks.
\end{itemize}

\subsection{Evaluation Metrics}
\label{appendix:eval_metric}

\paragraph{Success Rate (SR)} 
Following \citep{wu2024atlas}, we count a step as successful if the predicted action type and all parameters exactly match the ground truth. The success rate is the average accuracy under this criterion. Matching rules differ depending on the output format (see below).

\paragraph{Grounding (Grd.)} 
We adopt the criterion from \citep{wu2024atlas}, counting a prediction as correct if the relative distance between the predicted and ground-truth positions is less than 14\%.

\subsection{Qualitative Case Studies}
\label{appendix:case_studies}

\begin{figure*}[h]
\setlength{\abovecaptionskip}{0pt}
    \centering
    \begin{minipage}{0.72\textwidth}
        \begin{subfigure}[b]{0.3\textwidth}
            \centering
            \includegraphics[width=\textwidth]{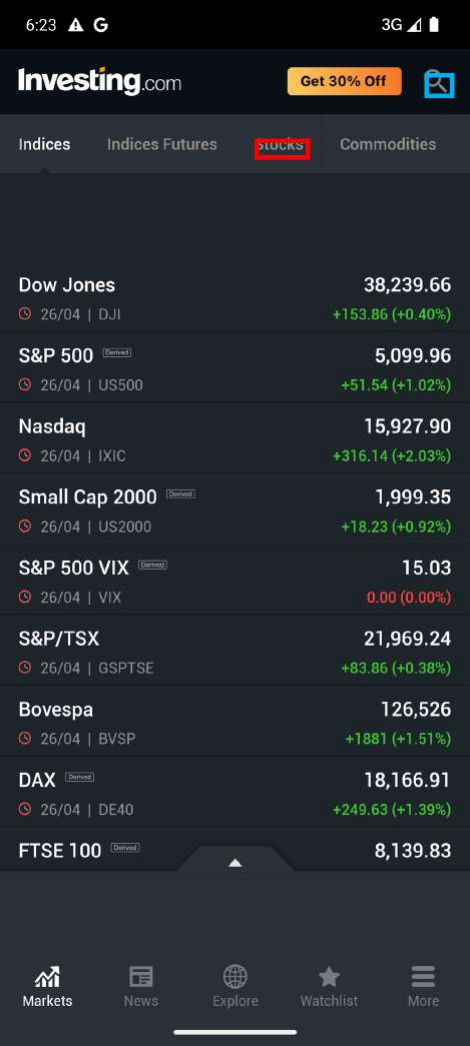}
            \caption{Search for the stock price trends of Google}
            \label{fig:procedural_effect}
        \end{subfigure}
        \hfill
        \begin{subfigure}[b]{0.325\textwidth}
            \centering
            \includegraphics[width=\textwidth]{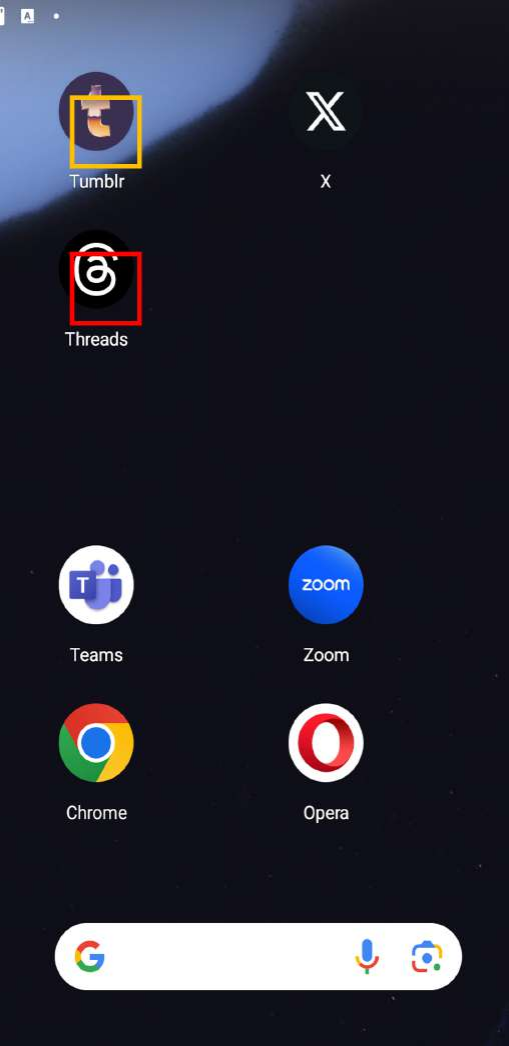}
            \caption{Use Tumblr to share the meeting link}
            \label{fig:stationary_effect}
        \end{subfigure}
        \hfill
        \begin{subfigure}[b]{0.3\textwidth}
            \centering
            \includegraphics[width=\textwidth]{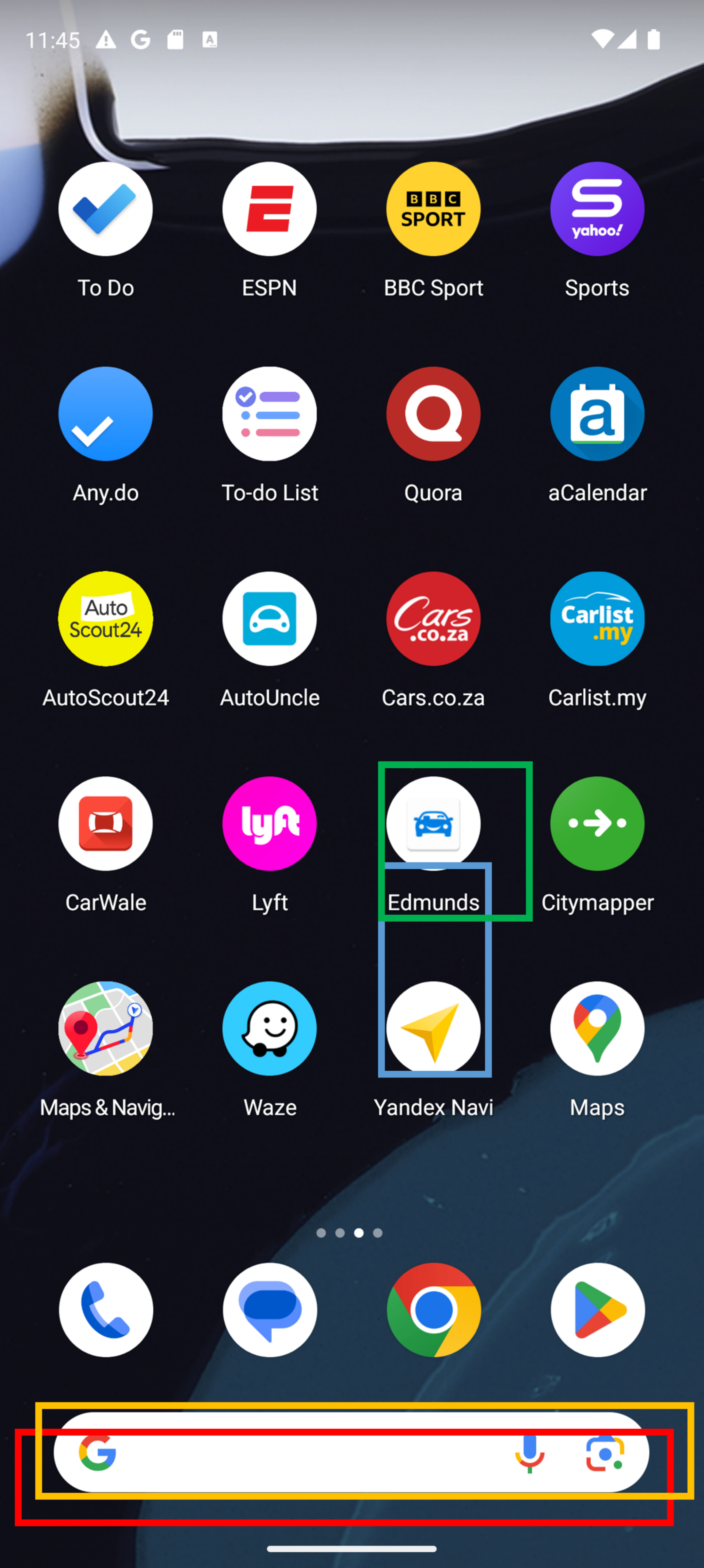}
            \caption{Verify the price of the leading product for Subaru with Edmunds}
            \label{fig:combined_effect}
        \end{subfigure}
    \end{minipage}
    \hfill
    \begin{minipage}{0.26\textwidth}
        \begin{subfigure}[b]{0.9\textwidth}
            \centering
            \includegraphics[width=0.5\textwidth]{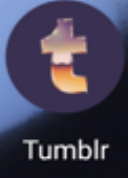}
            \caption{Icon retrieved from stationary memory in \ref{fig:stationary_effect}}
            \label{fig:retrieved_stationary}
        \end{subfigure}
        
        \vfill
        \begin{subfigure}[b]{0.9\textwidth}
            \centering
            \includegraphics[width=0.5\textwidth]{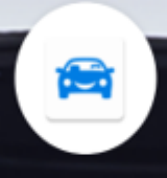}
            \caption{Icon retrieved by MAGNET in \ref{fig:combined_effect}}
            \label{fig:retrieved_magnet}
        \end{subfigure}
    \end{minipage}
    
    \caption{{Case studies on memory components.} {Red} boxes mark the \textcolor{red}{baseline}, {cyan} denote \textcolor{skyblue}{procedural memory}, {orange} denote \textcolor{orange}{stationary memory}, and {green} denote \textcolor{green}{\modelname}. Procedural (a) and stationary (b) memory both improve over the baseline, while \modelname~(c) outperforms them by combining both. (d) and (e) show the icon retrieval process for (b) and (c), respectively.}
    \label{fig:case_study}
    \vspace{1cm}
\end{figure*}

Figure~\ref{fig:case_study} presents three case studies showing how different components improve task execution.
\begin{enumerate}[leftmargin=0.5cm,itemindent=0cm,label=(\alph*)]
\item \textbf{Searching for Google stock.} The baseline clicks a misleading ``stack'' tab that only shows a stock leaderboard. With procedural memory, the agent recalls the correct workflow \texttt{Search for Stock Price Trends}, which includes the step \texttt{Tap the search icon and enter the company name}, thereby completing the task effectively.
\item \textbf{Launching Tumblr.} The baseline incorrectly identifies Tumblr as the Threads app. Stationary memory retrieves the exact Tumblr icon (\ref{fig:retrieved_stationary}), allowing the agent to select the correct app.
\item \textbf{Opening Edmunds for Subaru's leading product price verification.} Although the Subaru's leading model was already identified, the baseline ignores this and reverts to searching in the browser. Even with stationary memory, it still clicks the search bar. With procedural memory, the agent infers that the next step is to open Edmunds, but incorrectly selects Yandex Navi due to the absence of visual grounding. By combining both memories, \modelname plans the correct action and grounds it visually (\ref{fig:retrieved_magnet}), successfully opening Edmunds.
\end{enumerate}

\subsection{Detailed Architecture Ablation Results}
\label{appendix:detailed_ablation}

Table~\ref{tab:detailed_paired_ablation} presents the complete per-dataset results for the architecture ablation study summarized in Table~\ref{tab:paired_ablation} of the main paper.

\section{The Use of Large Language Models (LLMs)}

Large Language Models (LLMs) were used to aid in the writing and polishing of the manuscript. Specifically, we used an LLM to assist in refining the language, improving readability, and ensuring clarity in various sections of the paper. The model helped with tasks such as sentence rephrasing, grammar checking, and enhancing the overall flow of the text.

It is important to note that the LLM was not involved in the ideation, research methodology, or experimental design. All research concepts, ideas, and analyses were developed and conducted by the authors. The contributions of the LLM were solely focused on improving the linguistic quality of the paper, with no involvement in the scientific content or data analysis.

The authors take full responsibility for the content of the manuscript, including any text generated or polished by the LLM. We have ensured that the LLM-generated text adheres to ethical guidelines and does not contribute to plagiarism or scientific misconduct.
\input{cross_base_results}

\label{appendix:algorithms}
\newpage
\onecolumn
\section{Algorithm Details}

\begin{algorithm}[h]
\small
\caption{Instruction Clustering via Maximal Cliques}
\label{clustering_algorithm}
\begin{algorithmic}[1]
\REQUIRE Instructions $\{I_1, I_2, \ldots, I_n\}$, similarity threshold $\tau$ 
\ENSURE Instruction clusters $\mathcal{C} = \{c_1, c_2, \ldots, c_k\}$ 
\STATE Compute embeddings $e_i = \text{ENCODE}(I_i)$ for all instructions 
\STATE Initialize similarity graph $\mathcal{G} = (V, \emptyset)$ where $V = \{I_1, I_2, \ldots, I_n\}$ 
\FOR{each pair $(I_i, I_j)$ where $i \neq j$} 
    \IF{$\text{COSINE}(e_i, e_j) > \tau$} \STATE Add edge $(I_i, I_j)$ to $\mathcal{G}$ \ENDIF 
\ENDFOR 
\STATE $\mathcal{C} = \text{FIND\_MAXIMAL\_CLIQUES}(\mathcal{G})$ 
\RETURN $\mathcal{C}$ 
\end{algorithmic}
\end{algorithm}

\begin{algorithm}[h]
\small
\caption{Memory Retrieval and Update Mechanism}
\label{alg:memory_evolution}
\begin{algorithmic}[1]
\REQUIRE Query $x_q$ (natural language description of next action)
\REQUIRE Memory Bank $\mathcal{M} = \{m_1, m_2, \ldots, m_{|\mathcal{M}|}\}$ where each entry $m_i = \langle k_i, v_i, \tau_i, t_i, n_i \rangle$ contains:
\STATE \quad $k_i$: retrieval key (workflow description or UI function)
\STATE \quad $v_i$: stored value (workflow template or UI element patch)
\STATE \quad $\tau_i$: creation timestamp (when this entry was added to memory)
\STATE \quad $t_i$: last access counter (value of $C_{\text{global}}$ when last retrieved)
\STATE \quad $n_i$: total retrieval count (number of times this entry has been retrieved)
\REQUIRE Global retrieval counter $C_{\text{global}}$ (tracks total number of retrieval events)
\REQUIRE Parameters: $N$ (number of candidates to filter), $K$ (number of memories to retrieve)
\ENSURE Retrieved memories $\mathcal{M}_{\text{ret}}$, Updated $\mathcal{M}$ and $C_{\text{global}}$

\STATE \textbf{// Stage 1: Semantic Filtering}
\STATE Compute semantic similarity $s_i = \text{COSINE}(E(x_q), E(k_i))$ for all $m_i \in \mathcal{M}$
\STATE \quad where $E(\cdot)$ encodes text into embedding vectors
\STATE $\mathcal{C}_N = \text{TOP\_N}(\mathcal{M}, \text{key}=s_i)$ \COMMENT{Select top-N most semantically similar entries}

\STATE \textbf{// Stage 2: Dual-Key Ranking}
\FOR{each $m_i \in \mathcal{C}_N$}
    \STATE $g_i \leftarrow C_{\text{global}} - t_i$ \COMMENT{Retrieval gap: events since last access}
    \STATE $R_i \leftarrow \exp(-g_i / n_i)$ \COMMENT{retention score: usage-adaptive decay}
\ENDFOR

\STATE Sort $\mathcal{C}_N$ by $(R_i, \tau_i)$ in descending order 
\STATE \quad \COMMENT{Primary key: retention score $R_i$; Secondary key: creation time $\tau_i$}
\STATE $\mathcal{M}_{\text{ret}} = \text{TOP\_K}(\mathcal{C}_N)$ \COMMENT{Select top-K entries after sorting}

\STATE \textbf{// Stage 3: Update Retrieved Memories}
\STATE $C_{\text{global}} \leftarrow C_{\text{global}} + 1$ \COMMENT{Increment global counter}

\FOR{each $m_i \in \mathcal{M}_{\text{ret}}$}
    \STATE $t_i \leftarrow C_{\text{global}}$ \COMMENT{Record current retrieval event}
    \STATE $n_i \leftarrow n_i + 1$ \COMMENT{Increment retrieval count}
\ENDFOR

\IF{New workflow or UI element $\langle k^*, v^* \rangle$ extracted from successful task}
    \STATE $\tau^* \leftarrow \text{CURRENT\_TIMESTAMP}()$ \COMMENT{Record creation time}
    \STATE $\mathcal{M} \leftarrow \mathcal{M} \cup \{\langle k^*, v^*, \tau^*, C_{\text{global}}, 1 \rangle\}$ 
    \STATE \quad \COMMENT{Initialize: creation time $\tau^*$, last access $C_{\text{global}}$, count $1$}
\ENDIF

\RETURN $\mathcal{M}_{\text{ret}}$, $\mathcal{M}$, $C_{\text{global}}$
\end{algorithmic}
\end{algorithm}

\newpage
\section{Prompts}

\label{prompts}
\subsection{Memory Construction}
\label{memory_construction}


\begin{table}[H]
\begin{center}
\begin{tcolorbox}[colback=white!5!white,
                  colframe=gray!15!gray,
                  width=\textwidth,
                  title={Extract workflow for procedural memory.}]

You are an expert in analyzing and abstracting user behavior on mobile devices. Given a list of mobile tasks, each described by a natural language instruction and its corresponding action sequence, extract commonly used workflows shared across multiple tasks.

\textbf{Instructions:}
\begin{enumerate}[leftmargin=0.9cm]
    \item Identify repetitive subsequences of actions that appear in two or more tasks.
    \item Each extracted workflow should be a useful and reusable subroutine, not too specific to any one task.
    \item Do NOT output overlapping or highly similar workflows; each should be distinct and meaningful.
    \item Each workflow must contain at least 3 steps.
    \item Represent variable elements (such as user input, contact names, app names) using descriptive placeholders (e.g., \texttt{[SearchQuery]}, \texttt{[AppName]}, \texttt{[ContactName]}).
    \item Focus on semantic repetition, not just literal string match.
\end{enumerate}

\textbf{Workflow Examples:}
\begin{lstlisting}[language=json]
[
  {
    "title": "Install: Search and Install an App",
    "plan": [
      "Open the App Store or App Market or Play Store.",
      "Tap the search bar.",
      "Type [AppName] into the input field.",
      "Locate the correct app in the result list.",
      "Tap the [InstallButton] to download and install."
    ]
  },
  ...
]
\end{lstlisting}

\textbf{Input Mobile Tasks:}
\{tasks\}

\textbf{Output:}
Extract all valid reusable workflows from the input above, following the rules. Output strictly in JSON format as shown in the examples, without extra explanation or commentary.

\end{tcolorbox}
\end{center}
\end{table}

\begin{table}[H]
\begin{center}
\begin{tcolorbox}
[colback=white!5!white,colframe=gray!15!gray,width=\textwidth,title={Restate click action description for stationary memory.}]

Format the given context into: \texttt{click [ui element] to [purpose]}

\textbf{Requirements:}
\begin{enumerate}[leftmargin=0.9cm]
    \item \texttt{ui element}: concise name or description of the clicked UI element.
    \item \texttt{purpose}: clear explanation of the click's goal.
\end{enumerate}
\textbf{Examples:}
\begin{enumerate}[leftmargin=0.9cm]
    \item click the Chrome icon to search for information online
    \item click the More Info button to view quiz-related details
    \item click the second search result to read about hiking trails
\end{enumerate}

\textbf{Input Context:}

\quad Action Description: \{action description\}

\quad Action Result: \{action result\}

Output the formatted action description directly without any additional text.

\end{tcolorbox}
\end{center}    
\end{table}

\begin{table}[H]
\centering
\begin{tcolorbox}[
    colback=white!5!white,
    colframe=gray!15!gray,
    boxrule=1pt,
    width=\textwidth,
    title={Action Inference Prompt}
]
You are an assistant for inferring user actions based on mobile UI screenshots.

You will be given two screenshots:
\begin{enumerate}[leftmargin=0.9cm]
    \item \textbf{Before Action}: A screenshot of the UI before the user performed the action.
    \item \textbf{After Action}: A screenshot showing the result of the action.
\end{enumerate}

\textbf{Your task is to:}

\begin{tcolorbox}[
colback=white!5!white,colframe=gray!15!gray,boxrule=0.5pt, title={\textbf{General Action}}]
\begin{itemize}[leftmargin=*, partopsep=0pt, topsep=0pt]
    \setlength{\itemsep}{0pt}
    \setlength{\parsep}{0pt}
    \setlength{\parskip}{0pt}
    \item Provide a concise \textbf{action description} of what the user did. (e.g., "type 'blender recommendation' into the search bar.", "stop and set the task as completed/impossible."). Start with verb.
    \item Provide a concise \textbf{action result} describing the effect of the action.
\end{itemize}
\textbf{Output format (strictly JSON, no extra explanation or formatting):}
\begin{lstlisting}[language=json]
{
  "action_desc": "<description of the user action>",
  "action_result": "<description of the outcome of the action>"
}
\end{lstlisting}
\end{tcolorbox}

\begin{tcolorbox}[colback=white!5!white,colframe=gray!15!gray, boxrule=0.5pt, title={\textbf{Swipe Action}}]
\begin{itemize}[leftmargin=*, partopsep=0pt, topsep=0pt]
    \setlength{\itemsep}{0pt}
    \setlength{\parsep}{0pt}
    \setlength{\parskip}{0pt}
    \item Provide a concise \textbf{action description} of what the user did (e.g., "scroll up on the home feed").
    \item Provide a concise \textbf{action result} describing the effect of the action (e.g., "By doing this...").
\end{itemize}
\textbf{Output format (strictly JSON, no extra explanation or formatting):}
\begin{lstlisting}[language=json]
{
  "action_desc": "<a concise description of the user action. For swipe actions, only use direction terms: 'up', 'down', 'left', or 'right'>",
  "action_result": "<description of the outcome>"
}
\end{lstlisting}
\end{tcolorbox}

\begin{tcolorbox}[colback=white!5!white,colframe=gray!15!gray,boxrule=0.5pt, title={\textbf{Click Action}}]
\begin{itemize}[leftmargin=*, partopsep=0pt, topsep=0pt]
    \setlength{\itemsep}{0pt}
    \setlength{\parsep}{0pt}
    \setlength{\parskip}{0pt}
    \item Provide a concise \textbf{action description} of what the user clicked. Start with a verb. (e.g., "click on the settings icon in the top right")
    \item Provide a concise \textbf{action result} describing what happened after the click (e.g., "By doing this...").
\end{itemize}
\textbf{Output format (strictly JSON, no extra explanation or formatting):}
\begin{lstlisting}[language=json]
{
  "action_desc": "<description of the user action. (e.g., \"click on the settings icon in the top right\")>",
  "action_result": "<description of the outcome>"
}
\end{lstlisting}
\end{tcolorbox}

\textbf{Action Type} \{action\_type\}: \{action\_meaning\}
\label{tab:action_infer_prompt}
\end{tcolorbox}
\end{table}

\newpage
\subsection{GUI Agents}


\begin{table}[H]
\begin{center}
\begin{tcolorbox}
[colback=white!5!white,colframe=gray!15!gray,width=\textwidth,title={Observe: Generate screen description.}]

You are a smart and helpful visual assistant that is well-trained to describe smartphone screenshots.
\begin{enumerate}[leftmargin=0.5cm]
    \item You are provided with a screenshot of the current mobile phone.
    \item You are required to describe this screen's main content and its functionality. The output must be less than five sentences. 
    \item You are required to keep the description as concise and brief as possible.
\end{enumerate}

\textbf{Input:}

CURRENT SCREENSHOT: \{screenshot\}

YOUR RESPONSE:
      
\end{tcolorbox}
\end{center}
    \label{tab:observe_prompt}
\end{table}

\begin{table}[H]
\begin{center}
\begin{tcolorbox}
[colback=white!5!white,colframe=gray!15!gray,width=\textwidth,title={Plan: Generate next action description.}]

You are a smart and helpful visual assistant that is well-trained to manipulate mobile phones.
Your task is to navigate and take action on the current screen step-by-step to complete the user request.

\begin{enumerate}[leftmargin=1cm]
    \item You are provided with a screenshot of the current mobile phone, together with the textual screen description.
    \item You are provided with your history actions to decide on your next action. You can backtrack to revise the previous actions when necessary.
    \item You are provided with some relevant workflows for reference.
    \item You are required to analyze the task status and detail a reasonable future action plan to accomplish the user request.
\end{enumerate}

\textbf{Analysis Guidelines:}
\begin{enumerate}[leftmargin=1cm]
    \item You should check whether the historical actions have accomplished the user request.
    \item You should check the apps, icons, and buttons that are visible on the current screen and might pertain to the user request.
    \item You should combine the above information and describe your future action plan. If the given workflows are relevant and helpful, you may refer to them for guidance.
    \item The "Future Action Plan" must consist of a sequence of concrete, low-level action steps.
\end{enumerate}

\textbf{Output Format}
You are required to respond in a JSON format, consisting of 3 distinct parts with the following keys and corresponding content:

\begin{lstlisting}[language=json]
[
  {
    "Thought": "<Analyze the logic behind your next single-step action and your future action plan to fulfill the user request.>", 
    "Future Action Plan": [
      {"type": "<ACTION_TYPE>", "description": "<Natural language description of the action>"},
      ...
    ]
  }
]
\end{lstlisting}
CURRENT SCREENSHOT: \{screenshot\}

SCREEN DESCRIPTION: \{screen description\}

HISTORY ACTIONS: \{history actions\}

USER REQUEST: \{user request\}

RELEVANT WORKFLOWS: \{relevant workflows\}

YOUR RESPONSE:

\end{tcolorbox}
\end{center}
    \label{tab:plan_prompt}
\end{table}

\begin{table}[H]
\begin{center}
\begin{tcolorbox}
[colback=white!5!white,colframe=gray!15!gray,width=\textwidth,title={Predict: Generate detailed action arguments.}]

You are a smart and helpful visual assistant that is well-trained to manipulate mobile phones.  
Your task is to navigate on the current screen to complete the user request.  

\begin{enumerate}[leftmargin=1cm]
    \item You are provided with a screenshot of the current mobile phone.
    \item You are provided with a brief summarization of the screen content.
    \item You are provided with history actions trying to accomplish the user request, together with the previous action result that indicates how the current screenshot is obtained.
    \item You are provided with a \textbf{Relevant UI Element} that visually represents the most relevant UI component for the next action.
    \item You are required to decide on the next single-step valid action to be conducted on the current screen so as to fulfill the user request.
\end{enumerate}

\textbf{Valid Actions on the Screen:}  
\{action\_space\}  

\textbf{Output Format:}  
You must choose one of the valid APIs provided above and respond in the corresponding API call format.  
Your response should be strictly structured in JSON format, consisting of the following keys and corresponding content:  

\begin{lstlisting}[language=json]
{
  "THINK": "<Analyze the logic behind your next single-step action.>", 
  "NEXT": "<Describe the next single-step action in words, e.g. 'click on the .... located at ...'>", 
  "ACTION": "<Specify the precise API function name without arguments, e.g., click_element. Leave it empty if none applies or task is complete.>", 
  "ARGS": "<Specify arguments in dictionary format, e.g., {'bbox': [0,1,2,3]}. Leave empty if not needed or task complete.>", 
  "REASON": "<Explain your reasoning for choosing this action.>"
}
\end{lstlisting}

\textbf{Output Examples:}  

\begin{lstlisting}[language=json]
{
  "THINK": "...", 
  "NEXT": "...", 
  "ACTION": "click_element", 
  "ARGS": {"bbox": [100, 345, 219, 826]}
}
\end{lstlisting}

\begin{lstlisting}[language=json]
{
  "THINK": "...", 
  "NEXT": "...", 
  "ACTION": "scroll", 
  "ARGS": {"direction": "down"}
}
\end{lstlisting}

\begin{lstlisting}[language=json]
{
  "THINK": "...", 
  "NEXT": "...", 
  "ACTION": "press_home", 
  "ARGS": {}
}
\end{lstlisting}

\textbf{Inputs:}  

\begin{itemize}[leftmargin=*, partopsep=0pt, topsep=0pt]
    \setlength{\itemsep}{0pt}
    \setlength{\parsep}{0pt}
    \setlength{\parskip}{0pt}
    \item CURRENT SCREENSHOT: \{screenshot\}
    \item SCREEN CONTENT: \{screen\_desc\}
    \item HISTORY ACTIONS: \{history\_actions\}
    \item PREV ACTION RESULT: \{prev\_action\_result\}
    \item USER REQUEST: \{user\_request\}
    \item RELEVANT UI ELEMENT: \{relevant\_ui\}
    \item YOUR RESPONSE:
\end{itemize}

\end{tcolorbox}
\end{center}
    \label{tab:predict_prompt}
\end{table}

\begin{table}[H]
\begin{center}
\begin{tcolorbox}
[colback=white!5!white,colframe=gray!15!gray,width=\textwidth,title={Prompt for OS-Atlas.}]

You are a foundational action model capable of automating tasks across various digital environments, including desktop systems (Windows, macOS, Linux), mobile platforms (Android, iOS), and web browsers.  
You interact with digital devices in a human-like manner: by reading screenshots, analyzing them, and taking appropriate actions.

\textbf{Expertise:}
\begin{enumerate}
    \item \textbf{Grounding:} Given a screenshot and description, assist users in locating elements. Infer best-fit elements when implicit.
    \item \textbf{Executable Language Grounding:} Given a screenshot and task instruction, determine executable actions to complete the task.
\end{enumerate}

You are now operating in \textbf{Executable Language Grounding mode}. Your goal is to suggest executable actions that best fit user needs.

    
    
\textbf{1. Basic Actions (available across all platforms):}

\begin{lstlisting}[language=json]
[
    {
      "name": "CLICK",
      "purpose": "Click at the specified position.",
      "format": "CLICK <point>[[x-axis, y-axis]]</point>",
      "example": "CLICK <point>[[101, 872]]</point>"
    },
    {
      "name": "TYPE",
      "purpose": "Enter specified text at the designated location.",
      "format": "TYPE [input text]",
      "example": "TYPE [Shanghai shopping mall]"
    },
    {
      "name": "SCROLL",
      "purpose": "Scroll in the specified direction.",
      "format": "SCROLL [direction (UP/DOWN/LEFT/RIGHT)]",
      "example": "SCROLL [UP]"
    }
]
\end{lstlisting}

\textbf{2. Custom Actions (environment-specific):}  
Custom actions extend functionality to handle unseen or user-defined tasks.  

\{action\_space\}  

\textbf{Instruction:}  
In most cases, task instructions are high-level and abstract.  
Carefully read the instruction and action history, then reason to determine the most appropriate next action.  

\textbf{Output Requirements:}  
You must strictly generate two sections:
\begin{itemize}[leftmargin=*, partopsep=0pt, topsep=0pt]
    \setlength{\itemsep}{0pt}
    \setlength{\parsep}{0pt}
    \setlength{\parskip}{0pt}
    \item \texttt{Thoughts}: concise reasoning (max 20 words).
    \item \texttt{Actions}: the actual next one-step action.
\end{itemize}

\textbf{Inputs:}  
\begin{itemize}[leftmargin=*, partopsep=0pt, topsep=0pt]
    \setlength{\itemsep}{0pt}
    \setlength{\parsep}{0pt}
    \setlength{\parskip}{0pt}
    \item Screenshot: \{screenshot\}
    \item Task: \{user\_request\}
    \item History: \{history\_actions\}
\end{itemize}

\textbf{Output Format:}  

\begin{lstlisting}[language=json]
thoughts: "<a concise description of reasoning>"
actions: "<only next one step action usage>"
\end{lstlisting}

\end{tcolorbox}
\end{center}
    \caption{Prompt template for OS-Atlas model.}
    \label{tab:atlas_prompt}
\end{table}

\begin{table}[h]
\begin{center}
\begin{tcolorbox}
[colback=white!5!white,colframe=gray!15!gray,width=\textwidth,title={Action space for OS-Atlas.}]

\begin{lstlisting}[language=json]
[
    {
      "name": "PRESS_ENTER",
      "purpose": "Press an enter button to confirm the input, or submit the input, or start a new line of text.",
      "format": "PRESS_ENTER",
      "example": "PRESS_ENTER"
    },
    {
      "name": "PRESS_HOME",
      "purpose": "Press a home button to navigate to the home page.",
      "format": "PRESS_HOME",
      "example": "PRESS_HOME"
    },
    {
      "name": "PRESS_BACK",
      "purpose": "Press a back button to navigate to the previous screen.",
      "format": "PRESS_BACK",
      "example": "PRESS_BACK"
    },
    {
      "name": "STOP",
      "purpose": "Stop and set the state of the task.",
      "format": "STOP [task_status (SUCCESS/FAILURE)]",
      "example": "STOP [SUCCESS]"
    }
]
\end{lstlisting}

\end{tcolorbox}
\end{center}
    \label{tab:atlas_action_space_prompt}
\end{table}

\end{document}

%% file: main_result.tex
\begin{table*}[htbp]
\centering
\setlength{\abovecaptionskip}{2pt}
\resizebox{0.95\linewidth}{!}{
\setlength{\tabcolsep}{6pt} 

\begin{tabular}{lcccccc}
\toprule
\multirow{2}{*}{\textbf{Methods}} & \multicolumn{2}{c}{\textbf{AITZ}} & \multicolumn{2}{c}{\textbf{GUI-Odyssey}} & \multicolumn{2}{c}{\textbf{Amex}} \\
\cmidrule(lr){2-3} \cmidrule(lr){4-5} \cmidrule(lr){6-7}
& \textbf{SR (\%)} & \textbf{Grd. (\%)} & \textbf{SR (\%)} & \textbf{Grd. (\%)} & \textbf{SR (\%)} & \textbf{Grd. (\%)} \\
\midrule

\rowcolor{gray!15} 
\multicolumn{7}{c}{\textit{Specialized Models (Parameter-tuned)}} \\
Atlas-Pro-7B & \textbf{66.64} & 62.18 & 58.82 & 67.00 & \textbf{67.45} & 67.78 \\
GUI-R1-7B & 44.22 & 57.21 & 47.70 & 59.20 & 49.36 & 59.42 \\
UI-Venus-Navi-7B & 60.27 & \textbf{70.23} & \textbf{65.93} & \textbf{74.52} & 63.58 & \textbf{79.67} \\
InfiGUI-R1-3B & 49.49 & 55.34 & 55.51 & 64.98 & 62.00 & 69.98 \\

\midrule

\rowcolor{gray!15} 
\multicolumn{7}{c}{\textit{Agentic Frameworks (Training-free / Inference-only)}} \\
COAT (Qwen2.5-VL-32B) & 41.09  & 39.28  & 48.13  & 49.38  & 59.68  & 67.69  \\
Agent-S$^{\dagger}$ (Qwen2.5-VL-32B) & 42.98 & 42.87 & 49.21 & 50.74 & {58.29} & 69.85 \\
MAGNET$^{\dagger}$ (Qwen2.5-VL-32B) & 43.50  & 43.78  & \textbf{50.16}  & 51.91  & \textbf{62.84}  & 71.53  \\
MAGNET$^{\dagger}$ (Gemini-2.5-Pro) & \textbf{52.77} & \textbf{57.35} & {49.74} & \textbf{57.31} & {62.23} & \textbf{75.54} \\

\bottomrule
\end{tabular}

}

\caption{\textbf{Offline evaluation results.} We report Success Rate (SR) and Grounding Accuracy (Grd.). Methods are categorized into: (1) \textbf{Specialized Models} (parameter-tuned via SFT or RL) and (2) \textbf{Agentic Frameworks} (inference-only or retrieval-augmented based on frozen models). Methods marked with $\dagger$ denote memory-augmented frameworks. For agentic frameworks, \textbf{bold} indicates the best performance within the group.}
\label{tab:main_results}
\end{table*}

%% file: ablation_iid_new.tex
\begin{table}[t]
\centering
\small
\setlength{\abovecaptionskip}{2pt}
\setlength{\belowcaptionskip}{2pt}
\setlength{\tabcolsep}{2pt}

\begin{subtable}[t]{0.9\linewidth}
\centering
\begin{tabular*}{\linewidth}{@{\extracolsep{\fill}}cccccc}
\toprule[1.2pt]
\multicolumn{2}{c}{\textbf{Variants}} 
& \multicolumn{2}{c}{\textbf{GUI-Odyssey}} 
& \multicolumn{2}{c}{\textbf{Amex}} \\ 
\cmidrule(l){1-2} \cmidrule(l){3-4} \cmidrule(l){5-6}
\textbf{Stat.} & \textbf{Proc.} 
& \textbf{SR} & \textbf{Grd.} 
& \textbf{SR} & \textbf{Grd.} \\ 
\midrule
 &  & 48.13 & 49.38 & 59.68 & 67.69 \\ 
$\checkmark$ &  & 48.62 & 49.98 & 60.20 & 68.57 \\ 
 & $\checkmark$ & 49.72 & 51.28 & 62.34 & 70.86 \\ 
\midrule
$\checkmark$ & $\checkmark$ 
& \textbf{50.16} & \textbf{51.91} 
& \textbf{62.84} & \textbf{71.53} \\ 
\bottomrule[1.2pt]
\end{tabular*}
\caption{Results with Qwen2.5-VL-32B.}
\end{subtable}
\hfill
\begin{subtable}[t]{0.9\linewidth}
\centering
\begin{tabular*}{\linewidth}{@{\extracolsep{\fill}}cccccc}
\toprule[1.2pt]
\multicolumn{2}{c}{\textbf{Variants}} 
& \multicolumn{2}{c}{\textbf{GUI-Odyssey}} 
& \multicolumn{2}{c}{\textbf{Amex}} \\ 
\cmidrule(l){1-2} \cmidrule(l){3-4} \cmidrule(l){5-6}
\textbf{Stat.} & \textbf{Proc.} 
& \textbf{SR} & \textbf{Grd.} 
& \textbf{SR} & \textbf{Grd.} \\ 
\midrule
 &  & 48.92 & 55.95 & 60.35 & 73.34 \\ 
$\checkmark$ &  & 49.35 & 56.55 & 60.52 & 73.60 \\ 
 & $\checkmark$ & 49.65 & 57.18 & 62.11 & 75.40 \\ 
\midrule
$\checkmark$ & $\checkmark$ 
& \textbf{49.74} & \textbf{57.31} 
& \textbf{62.23} & \textbf{75.54} \\ 
\bottomrule[1.2pt]
\end{tabular*}
\caption{Results with Gemini-2.5-Pro.}
\label{tab:ablation_id_gemini}
\end{subtable}

\caption{\textbf{Ablation study on memory components of \modelname.} 
Results are reported on the ID subsets of GUI-Odyssey and Amex, where the same MLLM is used as both planner and actor. 
\textit{Stat.} and \textit{Proc.} denote stationary and procedural memory, respectively.}
\label{tab:ablation_iid}
\end{table}

%% file: cross_base_results_concise.tex
\begin{table}[t]
\centering
\small
\setlength{\tabcolsep}{6pt} 
\renewcommand{\arraystretch}{1.10}

\begin{tabular}{llcc}
\toprule
\multicolumn{2}{c}{\textbf{Backbones}} & \multicolumn{2}{c}{\textbf{Avg. Performance}} \\
\cmidrule(lr){1-2} \cmidrule(lr){3-4}
\textbf{Planner} & \textbf{Actor} & \textbf{$\Delta$SR (\%)} & \textbf{$\Delta$Grd. (\%)}  \\
\midrule
\cmidrule(lr){1-2} \cmidrule(lr){3-4}
\rowcolor{gray!8}
\multicolumn{4}{l}{\textit{Homogeneous Configurations}} \\
QwenVL & QwenVL & +2.5 & +3.6 \\
Gemini & Gemini & +1.5 & +2.3 \\
\rowcolor{gray!8}
\multicolumn{4}{l}{\textit{Heterogeneous Configurations}} \\
GPT-4o & OS-Atlas & +1.5 & +1.5 \\
QwenVL & OS-Atlas & +4.2 & +3.4 \\
Gemini & OS-Atlas & +1.7 & +2.6 \\
\midrule
\multicolumn{2}{c}{\textbf{Average}} & \textbf{+2.3} & \textbf{+2.7} \\
\bottomrule
\end{tabular}
\caption{\textbf{Effectiveness of MAGNET across architectures.} Average improvements ($\Delta$) with MAGNET across five planner-actor configurations, categorized into homogeneous (same model for planner and actor) and heterogeneous (different models). Results are averaged over AITZ, GUI-Odyssey, and Amex datasets. 
}
\label{tab:paired_ablation}
\end{table}

%% file: cross_base_results.tex
\begin{table*}[h!]
\centering
\small
\setlength{\tabcolsep}{4.5pt} 
\renewcommand{\arraystretch}{1.15}

\begin{tabular}{ll c cccccc}
\toprule
\multicolumn{2}{c}{\textbf{Backbones}} & \multirow{2}{*}{\textbf{MAG.}} & \multicolumn{2}{c}{\textbf{AITZ}} & \multicolumn{2}{c}{\textbf{GUI-Odyssey}} & \multicolumn{2}{c}{\textbf{Amex}} \\
\cmidrule(lr){1-2} \cmidrule(lr){4-5} \cmidrule(lr){6-7} \cmidrule(lr){8-9}
\multicolumn{1}{c}{\textbf{Planner}} & \multicolumn{1}{c}{\textbf{Actor}} & & \textbf{SR (\%)} & \textbf{Grd. (\%)} & \textbf{SR (\%)} & \textbf{Grd. (\%)} & \textbf{SR (\%)} & \textbf{Grd. (\%)} \\
\midrule

\rowcolor{gray!8} 
\multicolumn{9}{c}{\textit{Standard End-to-End Baselines}} \\
\multicolumn{2}{c}{Qwen2.5-VL-32B} & - & 40.62 & 38.71 & 47.54 & 48.86 & 59.02 & 66.93 \\
\multicolumn{2}{c}{Gemini-2.5-Pro} & - & 50.14 & 53.21 & 48.27 & 55.18 & 59.76 & 72.68 \\

\midrule

\rowcolor{gray!8} 
\multicolumn{9}{c}{\textit{Paired Backbone Ablation: Base vs. + MAGNET (Ours)}} \\

\multirow{2}{*}{GPT-4o} & \multirow{2}{*}{Atlas-Base-7B} 
& $\times$ & 36.92 & 36.54 & 33.84 & 35.30 & 42.17 & 50.16 \\
& & \checkmark & \textbf{38.38} & \textbf{38.55} & \textbf{36.20} & \textbf{37.64} & \textbf{42.80} & \textbf{50.32} \\
\cmidrule(lr){1-9}

\multirow{2}{*}{GPT-4o} & \multirow{2}{*}{Gemini-2.5-Pro} 
& $\times$ & 42.48 & 41.94 & 38.22 & 39.15 & 46.21 & 54.08 \\
& & \checkmark & \textbf{43.71} & \textbf{42.87} & \textbf{40.54} & \textbf{41.28} & \textbf{46.71} & \textbf{54.82} \\
\cmidrule(lr){1-9}

\multirow{2}{*}{GPT-4o} & \multirow{2}{*}{Qwen2.5-VL-32B} 
& $\times$ & 34.37 & 34.29 & 32.04 & 33.10 & 39.48 & 47.26 \\
& & \checkmark & \textbf{36.15} & \textbf{35.82} & \textbf{34.29} & \textbf{34.64} & \textbf{41.72} & \textbf{49.28} \\
\cmidrule(lr){1-9}

\multirow{2}{*}{Qwen2.5-VL-32B} & \multirow{2}{*}{Atlas-Base-7B} 
& $\times$ & 43.64 & 41.53 & 49.93 & 51.58 & 53.92 & 70.59 \\
& & \checkmark & \textbf{45.73} & \textbf{46.51} & \textbf{52.07} & \textbf{54.91} & \textbf{62.37} & \textbf{72.57} \\
\cmidrule(lr){1-9}

\multirow{2}{*}{Qwen2.5-VL-32B} & \multirow{2}{*}{Qwen2.5-VL-32B} 
& $\times$ & 41.09 & 39.28 & 48.13 & 49.38 & 59.68 & 67.69 \\
& & \checkmark & \textbf{43.50} & \textbf{43.78} & \textbf{50.16} & \textbf{51.91} & \textbf{62.84} & \textbf{71.53} \\
\cmidrule(lr){1-9}

\multirow{2}{*}{Gemini-2.5-Pro} & \multirow{2}{*}{Atlas-Base-7B} 
& $\times$ & 45.31 & 48.48 & 44.54 & 52.10 & 56.31 & 69.42 \\
& & \checkmark & \textbf{47.44} & \textbf{53.03} & \textbf{45.40} & \textbf{53.67} & \textbf{58.32} & \textbf{71.04} \\
\cmidrule(lr){1-9}

\multirow{2}{*}{Gemini-2.5-Pro} & \multirow{2}{*}{Gemini-2.5-Pro} 
& $\times$ & 50.87 & 53.88 & 48.92 & 55.95 & 60.35 & 73.34 \\
& & \checkmark & \textbf{52.77} & \textbf{57.35} & \textbf{49.74} & \textbf{57.31} & \textbf{62.23} & \textbf{75.54} \\

\bottomrule
\end{tabular}
\caption{Backbone-agnostic effectiveness of MAGNET. By grouping backbones in pairs, this layout highlights the consistent performance gain across different LLM/VLM combinations. \textbf{Bold} indicates the better performance within each backbone pair.}
\label{tab:detailed_paired_ablation}
\end{table*}

%% file: custom.bib
@inproceedings{park2023generative,
  title={Generative agents: Interactive simulacra of human behavior},
  author={Park, Joon Sung and O'Brien, Joseph and Cai, Carrie Jun and Morris, Meredith Ringel and Liang, Percy and Bernstein, Michael S},
  booktitle={Proceedings of the 36th annual acm symposium on user interface software and technology},
  pages={1--22},
  year={2023}
}

@article{liu2024autoglm,
  title={Autoglm: Autonomous foundation agents for guis},
  author={Liu, Xiao and Qin, Bo and Liang, Dongzhu and Dong, Guang and Lai, Hanyu and Zhang, Hanchen and Zhao, Hanlin and Iong, Iat Long and Sun, Jiadai and Wang, Jiaqi and others},
  journal={arXiv preprint arXiv:2411.00820},
  year={2024}
}

@article{wu2024atlas,
  title={Os-atlas: A foundation action model for generalist gui agents},
  author={Wu, Zhiyong and Wu, Zhenyu and Xu, Fangzhi and Wang, Yian and Sun, Qiushi and Jia, Chengyou and Cheng, Kanzhi and Ding, Zichen and Chen, Liheng and Liang, Paul Pu and others},
  journal={arXiv preprint arXiv:2410.23218},
  year={2024}
}

@article{wu2024copilot,
  title={Os-copilot: Towards generalist computer agents with self-improvement},
  author={Wu, Zhiyong and Han, Chengcheng and Ding, Zichen and Weng, Zhenmin and Liu, Zhoumianze and Yao, Shunyu and Yu, Tao and Kong, Lingpeng},
  journal={arXiv preprint arXiv:2402.07456},
  year={2024}
}

@article{gao2025chain,
  title={Chain-of-Memory: Enhancing GUI Agents for Cross-Application Navigation},
  author={Gao, Xinzge and Hu, Chuanrui and Chen, Bin and Li, Teng},
  journal={arXiv preprint arXiv:2506.18158},
  year={2025}
}

@article{hu2025agents,
  title={Os agents: A survey on mllm-based agents for general computing devices use},
  author={Hu, Xueyu and Xiong, Tao and Yi, Biao and Wei, Zishu and Xiao, Ruixuan and Chen, Yurun and Ye, Jiasheng and Tao, Meiling and Zhou, Xiangxin and Zhao, Ziyu and others},
  journal={arXiv preprint arXiv:2508.04482},
  year={2025}
}

@article{jiang2025appagentx,
  title={Appagentx: Evolving gui agents as proficient smartphone users},
  author={Jiang, Wenjia and Zhuang, Yangyang and Song, Chenxi and Yang, Xu and Zhou, Joey Tianyi and Zhang, Chi},
  journal={arXiv preprint arXiv:2503.02268},
  year={2025}
}

@inproceedings{zhong2024memorybank,
  title={Memorybank: Enhancing large language models with long-term memory},
  author={Zhong, Wanjun and Guo, Lianghong and Gao, Qiqi and Ye, He and Wang, Yanlin},
  booktitle={Proceedings of the AAAI Conference on Artificial Intelligence},
  volume={38},
  number={17},
  pages={19724--19731},
  year={2024}
}

@article{hu2024hiagent,
  title={Hiagent: Hierarchical working memory management for solving long-horizon agent tasks with large language model},
  author={Hu, Mengkang and Chen, Tianxing and Chen, Qiguang and Mu, Yao and Shao, Wenqi and Luo, Ping},
  journal={arXiv preprint arXiv:2408.09559},
  year={2024}
}

@article{wang2024symbolic,
  title={Symbolic working memory enhances language models for complex rule application},
  author={Wang, Siyuan and Wei, Zhongyu and Choi, Yejin and Ren, Xiang},
  journal={arXiv preprint arXiv:2408.13654},
  year={2024}
}

@article{wang2023voyager,
  title={Voyager: An open-ended embodied agent with large language models},
  author={Wang, Guanzhi and Xie, Yuqi and Jiang, Yunfan and Mandlekar, Ajay and Xiao, Chaowei and Zhu, Yuke and Fan, Linxi and Anandkumar, Anima},
  journal={arXiv preprint arXiv:2305.16291},
  year={2023}
}

@article{yang2024buffer,
  title={Buffer of thoughts: Thought-augmented reasoning with large language models},
  author={Yang, Ling and Yu, Zhaochen and Zhang, Tianjun and Cao, Shiyi and Xu, Minkai and Zhang, Wentao and Gonzalez, Joseph E and Cui, Bin},
  journal={Advances in Neural Information Processing Systems},
  volume={37},
  pages={113519--113544},
  year={2024}
}

@inproceedings{hong2024cogagent,
  title={Cogagent: A visual language model for gui agents},
  author={Hong, Wenyi and Wang, Weihan and Lv, Qingsong and Xu, Jiazheng and Yu, Wenmeng and Ji, Junhui and Wang, Yan and Wang, Zihan and Dong, Yuxiao and Ding, Ming and others},
  booktitle={Proceedings of the IEEE/CVF Conference on Computer Vision and Pattern Recognition},
  pages={14281--14290},
  year={2024}
}

@inproceedings{zhang2025appagent,
  title={Appagent: Multimodal agents as smartphone users},
  author={Zhang, Chi and Yang, Zhao and Liu, Jiaxuan and Li, Yanda and Han, Yucheng and Chen, Xin and Huang, Zebiao and Fu, Bin and Yu, Gang},
  booktitle={Proceedings of the 2025 CHI Conference on Human Factors in Computing Systems},
  pages={1--20},
  year={2025}
}

@article{cheng2024seeclick,
  title={Seeclick: Harnessing gui grounding for advanced visual gui agents},
  author={Cheng, Kanzhi and Sun, Qiushi and Chu, Yougang and Xu, Fangzhi and Li, Yantao and Zhang, Jianbing and Wu, Zhiyong},
  journal={arXiv preprint arXiv:2401.10935},
  year={2024}
}

@article{wang2024genartist,
  title={Genartist: Multimodal llm as an agent for unified image generation and editing},
  author={Wang, Zhenyu and Li, Aoxue and Li, Zhenguo and Liu, Xihui},
  journal={Advances in Neural Information Processing Systems},
  volume={37},
  pages={128374--128395},
  year={2024}
}

@article{wu2023visual,
  title={Visual chatgpt: Talking, drawing and editing with visual foundation models},
  author={Wu, Chenfei and Yin, Shengming and Qi, Weizhen and Wang, Xiaodong and Tang, Zecheng and Duan, Nan},
  journal={arXiv preprint arXiv:2303.04671},
  year={2023}
}

@article{yang2023mm,
  title={Mm-react: Prompting chatgpt for multimodal reasoning and action},
  author={Yang, Zhengyuan and Li, Linjie and Wang, Jianfeng and Lin, Kevin and Azarnasab, Ehsan and Ahmed, Faisal and Liu, Zicheng and Liu, Ce and Zeng, Michael and Wang, Lijuan},
  journal={arXiv preprint arXiv:2303.11381},
  year={2023}
}

@article{wang2024mobile,
  title={Mobile-agent: Autonomous multi-modal mobile device agent with visual perception},
  author={Wang, Junyang and Xu, Haiyang and Ye, Jiabo and Yan, Ming and Shen, Weizhou and Zhang, Ji and Huang, Fei and Sang, Jitao},
  journal={arXiv preprint arXiv:2401.16158},
  year={2024}
}

@inproceedings{wang2025mllm,
  title={Mllm-tool: A multimodal large language model for tool agent learning},
  author={Wang, Chenyu and Luo, Weixin and Dong, Sixun and Xuan, Xiaohua and Li, Zhengxin and Ma, Lin and Gao, Shenghua},
  booktitle={2025 IEEE/CVF Winter Conference on Applications of Computer Vision (WACV)},
  pages={6678--6687},
  year={2025},
  organization={IEEE}
}

@article{xu2025mem,
  title={A-mem: Agentic memory for llm agents},
  author={Xu, Wujiang and Mei, Kai and Gao, Hang and Tan, Juntao and Liang, Zujie and Zhang, Yongfeng},
  journal={arXiv preprint arXiv:2502.12110},
  year={2025}
}

@article{modarressi2023ret,
  title={Ret-llm: Towards a general read-write memory for large language models},
  author={Modarressi, Ali and Imani, Ayyoob and Fayyaz, Mohsen and Sch{\"u}tze, Hinrich},
  journal={arXiv preprint arXiv:2305.14322},
  year={2023}
}

@article{packer2023memgpt,
  title={MemGPT: Towards LLMs as Operating Systems.},
  author={Packer, Charles and Fang, Vivian and Patil, Shishir\_G and Lin, Kevin and Wooders, Sarah and Gonzalez, Joseph\_E},
  year={2023},
  publisher={ArXiv}
}

@article{bai2025qwen2,
  title={Qwen2. 5-vl technical report},
  author={Bai, Shuai and Chen, Keqin and Liu, Xuejing and Wang, Jialin and Ge, Wenbin and Song, Sibo and Dang, Kai and Wang, Peng and Wang, Shijie and Tang, Jun and others},
  journal={arXiv preprint arXiv:2502.13923},
  year={2025}
}

@article{lu2024gui,
  title={Gui odyssey: A comprehensive dataset for cross-app gui navigation on mobile devices},
  author={Lu, Quanfeng and Shao, Wenqi and Liu, Zitao and Meng, Fanqing and Li, Boxuan and Chen, Botong and Huang, Siyuan and Zhang, Kaipeng and Qiao, Yu and Luo, Ping},
  journal={arXiv preprint arXiv:2406.08451},
  year={2024}
}

@article{chai2024amex,
  title={Amex: Android multi-annotation expo dataset for mobile gui agents},
  author={Chai, Yuxiang and Huang, Siyuan and Niu, Yazhe and Xiao, Han and Liu, Liang and Zhang, Dingyu and Ren, Shuai and Li, Hongsheng},
  journal={arXiv preprint arXiv:2407.17490},
  year={2024}
}

@article{zhang2024android,
  title={Android in the zoo: Chain-of-action-thought for gui agents},
  author={Zhang, Jiwen and Wu, Jihao and Teng, Yihua and Liao, Minghui and Xu, Nuo and Xiao, Xiao and Wei, Zhongyu and Tang, Duyu},
  journal={arXiv preprint arXiv:2403.02713},
  year={2024}
}

@article{xu2024aguvis,
  title={Aguvis: Unified pure vision agents for autonomous gui interaction},
  author={Xu, Yiheng and Wang, Zekun and Wang, Junli and Lu, Dunjie and Xie, Tianbao and Saha, Amrita and Sahoo, Doyen and Yu, Tao and Xiong, Caiming},
  journal={arXiv preprint arXiv:2412.04454},
  year={2024}
}

@article{li2023modelscope,
  title={Modelscope-agent: Building your customizable agent system with open-source large language models},
  author={Li, Chenliang and Chen, Hehong and Yan, Ming and Shen, Weizhou and Xu, Haiyang and Wu, Zhikai and Zhang, Zhicheng and Zhou, Wenmeng and Chen, Yingda and Cheng, Chen and others},
  journal={arXiv preprint arXiv:2309.00986},
  year={2023}
}

@article{shen2024small,
  title={Small llms are weak tool learners: A multi-llm agent},
  author={Shen, Weizhou and Li, Chenliang and Chen, Hongzhan and Yan, Ming and Quan, Xiaojun and Chen, Hehong and Zhang, Ji and Huang, Fei},
  journal={arXiv preprint arXiv:2401.07324},
  year={2024}
}

@article{schick2023toolformer,
  title={Toolformer: Language models can teach themselves to use tools},
  author={Schick, Timo and Dwivedi-Yu, Jane and Dess{\`\i}, Roberto and Raileanu, Roberta and Lomeli, Maria and Hambro, Eric and Zettlemoyer, Luke and Cancedda, Nicola and Scialom, Thomas},
  journal={Advances in Neural Information Processing Systems},
  volume={36},
  pages={68539--68551},
  year={2023}
}

@article{qian2024iterative,
  title={Iterative experience refinement of software-developing agents},
  author={Qian, Chen and Li, Jiahao and Dang, Yufan and Liu, Wei and Wang, YiFei and Xie, Zihao and Chen, Weize and Yang, Cheng and Zhang, Yingli and Liu, Zhiyuan and others},
  journal={arXiv preprint arXiv:2405.04219},
  year={2024}
}

@article{xi2024agentgym,
  title={Agentgym: Evolving large language model-based agents across diverse environments},
  author={Xi, Zhiheng and Ding, Yiwen and Chen, Wenxiang and Hong, Boyang and Guo, Honglin and Wang, Junzhe and Yang, Dingwen and Liao, Chenyang and Guo, Xin and He, Wei and others},
  journal={arXiv preprint arXiv:2406.04151},
  year={2024}
}

@article{li2025optimus,
  title={Optimus-3: Towards Generalist Multimodal Minecraft Agents with Scalable Task Experts},
  author={Li, Zaijing and Xie, Yuquan and Shao, Rui and Chen, Gongwei and Guan, Weili and Jiang, Dongmei and Nie, Liqiang},
  journal={arXiv preprint arXiv:2506.10357},
  year={2025}
}

@article{li2024mmedagent,
  title={Mmedagent: Learning to use medical tools with multi-modal agent},
  author={Li, Binxu and Yan, Tiankai and Pan, Yuanting and Luo, Jie and Ji, Ruiyang and Ding, Jiayuan and Xu, Zhe and Liu, Shilong and Dong, Haoyu and Lin, Zihao and others},
  journal={arXiv preprint arXiv:2407.02483},
  year={2024}
}

@article{gong2025mindflow,
  title={Mindflow: Revolutionizing e-commerce customer support with multimodal llm agents},
  author={Gong, Ming and Huang, Xucheng and Yang, Chenghan and Peng, Xianhan and Wang, Haoxin and Liu, Yang and Jiang, Ling},
  journal={arXiv preprint arXiv:2507.05330},
  year={2025}
}

@article{li2408appagent,
  title={Appagent v2: Advanced agent for flexible mobile interactions, 2024a},
  author={Li, Yanda and Zhang, Chi and Yang, Wanqi and Fu, Bin and Cheng, Pei and Chen, Xin and Chen, Ling and Wei, Yunchao},
  journal={URL https://arxiv. org/abs/2408.11824},
  volume={2}
}

@article{deng2023mind2web,
  title={Mind2web: Towards a generalist agent for the web},
  author={Deng, Xiang and Gu, Yu and Zheng, Boyuan and Chen, Shijie and Stevens, Sam and Wang, Boshi and Sun, Huan and Su, Yu},
  journal={Advances in Neural Information Processing Systems},
  volume={36},
  pages={28091--28114},
  year={2023}
}

@article{zhang2024ui,
  title={Ui-hawk: Unleashing the screen stream understanding for gui agents},
  author={Zhang, Jiwen and Yu, Yaqi and Liao, Minghui and Li, Wentao and Wu, Jihao and Wei, Zhongyu},
  year={2024},
  publisher={Preprints}
}

@article{luo2025gui,
  title={Gui-r1: A generalist r1-style vision-language action model for gui agents},
  author={Luo, Run and Wang, Lu and He, Wanwei and Xia, Xiaobo},
  journal={arXiv preprint arXiv:2504.10458},
  year={2025}
}

@article{wang2024mobilev2,
  title={Mobile-agent-v2: Mobile device operation assistant with effective navigation via multi-agent collaboration},
  author={Wang, Junyang and Xu, Haiyang and Jia, Haitao and Zhang, Xi and Yan, Ming and Shen, Weizhou and Zhang, Ji and Huang, Fei and Sang, Jitao},
  journal={Advances in Neural Information Processing Systems},
  volume={37},
  pages={2686--2710},
  year={2024}
}

@article{wang2025mobile,
  title={Mobile-agent-e: Self-evolving mobile assistant for complex tasks},
  author={Wang, Zhenhailong and Xu, Haiyang and Wang, Junyang and Zhang, Xi and Yan, Ming and Zhang, Ji and Huang, Fei and Ji, Heng},
  journal={arXiv preprint arXiv:2501.11733},
  year={2025}
}

@article{qin2025ui,
  title={Ui-tars: Pioneering automated gui interaction with native agents},
  author={Qin, Yujia and Ye, Yining and Fang, Junjie and Wang, Haoming and Liang, Shihao and Tian, Shizuo and Zhang, Junda and Li, Jiahao and Li, Yunxin and Huang, Shijue and others},
  journal={arXiv preprint arXiv:2501.12326},
  year={2025}
}

@inproceedings{chen2025guicourse,
  title={GUICourse: From General Vision Language Model to Versatile GUI Agent},
  author={Chen, Wentong and Cui, Junbo and Hu, Jinyi and Qin, Yujia and Fang, Junjie and Zhao, Yue and Wang, Chongyi and Liu, Jun and Chen, Guirong and Huo, Yupeng and others},
  booktitle={Proceedings of the 63rd Annual Meeting of the Association for Computational Linguistics (Volume 1: Long Papers)},
  pages={21936--21959},
  year={2025}
}

@misc{google_gemini_2_5,
    author = {{Gemini Team, Google}},
    title = {Gemini-2.5-Pro},
    howpublished = {\url{https://deepmind.google/models/gemini/pro/}},
    year = {2025},
    month = {June},
    version = {2.5}
}

@misc{openai_gpt4o,
    author = {OpenAI},
    title = {GPT-4O},
    howpublished = {\url{https://openai.com/index/hello-gpt-4o/}},
    year = {2024},
    month = {May},
}

@article{gou2024navigating,
  title={Navigating the digital world as humans do: Universal visual grounding for gui agents},
  author={Gou, Boyu and Wang, Ruohan and Zheng, Boyuan and Xie, Yanan and Chang, Cheng and Shu, Yiheng and Sun, Huan and Su, Yu},
  journal={arXiv preprint arXiv:2410.05243},
  year={2024}
}

@article{wang2024jarvis,
  title={Jarvis-1: Open-world multi-task agents with memory-augmented multimodal language models},
  author={Wang, Zihao and Cai, Shaofei and Liu, Anji and Jin, Yonggang and Hou, Jinbing and Zhang, Bowei and Lin, Haowei and He, Zhaofeng and Zheng, Zilong and Yang, Yaodong and others},
  journal={IEEE Transactions on Pattern Analysis and Machine Intelligence},
  year={2024},
  publisher={IEEE}
}

@inproceedings{fan2024videoagent,
  title={Videoagent: A memory-augmented multimodal agent for video understanding},
  author={Fan, Yue and Ma, Xiaojian and Wu, Rujie and Du, Yuntao and Li, Jiaqi and Gao, Zhi and Li, Qing},
  booktitle={European Conference on Computer Vision},
  pages={75--92},
  year={2024},
  organization={Springer}
}

@article{sun2019using,
  title={Using statistical measures and machine learning for graph reduction to solve maximum weight clique problems},
  author={Sun, Yuan and Li, Xiaodong and Ernst, Andreas},
  journal={IEEE transactions on pattern analysis and machine intelligence},
  volume={43},
  number={5},
  pages={1746--1760},
  year={2019},
  publisher={IEEE}
}

@article{yu2025omniparser,
  title={Omniparser v2: Structured-points-of-thought for unified visual text parsing and its generality to multimodal large language models},
  author={Yu, Wenwen and Yang, Zhibo and Wan, Jianqiang and Song, Sibo and Tang, Jun and Cheng, Wenqing and Liu, Yuliang and Bai, Xiang},
  journal={arXiv preprint arXiv:2502.16161},
  year={2025}
}

@article{wang2025ui,
  title={Ui-tars-2 technical report: Advancing gui agent with multi-turn reinforcement learning},
  author={Wang, Haoming and Zou, Haoyang and Song, Huatong and Feng, Jiazhan and Fang, Junjie and Lu, Junting and Liu, Longxiang and Luo, Qinyu and Liang, Shihao and Huang, Shijue and others},
  journal={arXiv preprint arXiv:2509.02544},
  year={2025}
}

@article{agashe2024agent,
  title={Agent s: An open agentic framework that uses computers like a human},
  author={Agashe, Saaket and Han, Jiuzhou and Gan, Shuyu and Yang, Jiachen and Li, Ang and Wang, Xin Eric},
  journal={arXiv preprint arXiv:2410.08164},
  year={2024}
}

@article{rawles2024androidworld,
  title={Androidworld: A dynamic benchmarking environment for autonomous agents},
  author={Rawles, Christopher and Clinckemaillie, Sarah and Chang, Yifan and Waltz, Jonathan and Lau, Gabrielle and Fair, Marybeth and Li, Alice and Bishop, William and Li, Wei and Campbell-Ajala, Folawiyo and others},
  journal={arXiv preprint arXiv:2405.14573},
  year={2024}
}

@article{hu2024minicpm,
  title={Minicpm: Unveiling the potential of small language models with scalable training strategies},
  author={Hu, Shengding and Tu, Yuge and Han, Xu and He, Chaoqun and Cui, Ganqu and Long, Xiang and Zheng, Zhi and Fang, Yewei and Huang, Yuxiang and Zhao, Weilin and others},
  journal={arXiv preprint arXiv:2404.06395},
  year={2024}
}

@book{ebbinghaus1885gedachtnis,
  title={{\"U}ber das ged{\"a}chtnis: untersuchungen zur experimentellen psychologie},
  author={Ebbinghaus, Hermann},
  year={1885},
  publisher={Duncker \& Humblot}
}

@article{gu2025ui,
  title={Ui-venus technical report: Building high-performance ui agents with rft},
  author={Gu, Zhangxuan and Zeng, Zhengwen and Xu, Zhenyu and Zhou, Xingran and Shen, Shuheng and Liu, Yunfei and Zhou, Beitong and Meng, Changhua and Xia, Tianyu and Chen, Weizhi and others},
  journal={arXiv preprint arXiv:2508.10833},
  year={2025}
}

@article{liu2025infigui,
  title={Infigui-r1: Advancing multimodal gui agents from reactive actors to deliberative reasoners},
  author={Liu, Yuhang and Li, Pengxiang and Xie, Congkai and Hu, Xavier and Han, Xiaotian and Zhang, Shengyu and Yang, Hongxia and Wu, Fei},
  journal={arXiv preprint arXiv:2504.14239},
  year={2025}
}

@book{shneiderman2010designing,
  title={Designing the user interface: strategies for effective human-computer interaction},
  author={Shneiderman, Ben},
  year={2010},
  publisher={Pearson Education India}
}

@article{sheridan2005human,
  title={Human-automation interaction},
  author={Sheridan, Thomas B and Parasuraman, Raja},
  journal={Reviews of human factors and ergonomics},
  volume={1},
  number={1},
  pages={89--129},
  year={2005},
  publisher={SAGE Publications Sage CA: Los Angeles, CA}
}

@incollection{hutchins1986direct,
  title={Direct manipulation interfaces},
  author={Hutchins, Edwin L and Hollan, James D and Norman, Donald A},
  booktitle={User centered system design},
  pages={87--124},
  year={1986},
  publisher={CRC Press}
}

@article{yang2025gui,
  title={GUI-Robust: A Comprehensive Dataset for Testing GUI Agent Robustness in Real-World Anomalies},
  author={Yang, Jingqi and Song, Zhilong and Chen, Jiawei and Song, Mingli and Zhou, Sheng and Ouyang, Xiaogang and Chen, Chun and Wang, Can and others},
  journal={arXiv preprint arXiv:2506.14477},
  year={2025}
}

@article{wang2024comprehensive,
  title={A comprehensive survey of continual learning: Theory, method and application},
  author={Wang, Liyuan and Zhang, Xingxing and Su, Hang and Zhu, Jun},
  journal={IEEE transactions on pattern analysis and machine intelligence},
  volume={46},
  number={8},
  pages={5362--5383},
  year={2024},
  publisher={IEEE}
}

@article{sager2025comprehensive,
  title={A Comprehensive Survey of Agents for Computer Use: Foundations, Challenges, and Future Directions},
  author={Sager, Pascal J and Meyer, Benjamin and Yan, Peng and von Wartburg-Kottler, Rebekka and Etaiwi, Layan and Enayati, Aref and Nobel, Gabriel and Abdulkadir, Ahmed and Grewe, Benjamin F and Stadelmann, Thilo},
  journal={arXiv preprint arXiv:2501.16150},
  year={2025}
}

@article{zheng2024gpt,
  title={Gpt-4v (ision) is a generalist web agent, if grounded},
  author={Zheng, Boyuan and Gou, Boyu and Kil, Jihyung and Sun, Huan and Su, Yu},
  journal={arXiv preprint arXiv:2401.01614},
  year={2024}
}
